\documentclass[manuscript,sigconf]{acmart}

\AtBeginDocument{%
  \providecommand\BibTeX{{%
    \normalfont B\kern-0.5em{\scshape i\kern-0.25em b}\kern-0.8em\TeX}}}

\copyrightyear{2023}
\acmYear{2023}
\setcopyright{acmlicensed}\acmConference[AIES '23]{AAAI/ACM Conference on AI, Ethics, and Society}{August 8--10, 2023}{Montréal, QC, Canada}
\acmBooktitle{AAAI/ACM Conference on AI, Ethics, and Society (AIES '23), August 8--10, 2023, Montréal, QC, Canada}
\acmPrice{15.00}
\acmDOI{10.1145/3600211.3604710}
\acmISBN{979-8-4007-0231-0/23/08}

\usepackage[english]{babel}
\usepackage{bbm}
\usepackage{amsmath, amsthm}
\usepackage{graphicx}
\usepackage{amsmath}

\usepackage{amssymb}

\usepackage{amsmath}
\usepackage{graphicx}

\usepackage{outlines}

\usepackage{wrapfig}
\usepackage[ruled,vlined]{algorithm2e}
\usepackage{cleveref}
\crefformat{section}{\S#2#1#3}
\usepackage{bm}

\usepackage{lipsum}                     

\usepackage{subcaption}

\usepackage{enumitem}
\setlist[enumerate]{nosep}
\setlist[itemize]{nosep}
\makeatletter 
\g@addto@macro\normalsize{
  \setlength\abovedisplayskip{1pt}
  \setlength\belowdisplayskip{1pt}
  \setlength\abovedisplayshortskip{1pt}
  \setlength\belowdisplayshortskip{1pt}
}
\makeatother
\usepackage[font=small,labelfont=bf]{caption}
\newtheoremstyle{mainstyle}
  {4pt} 
  {4pt} 
  {\itshape} 
  {} 
  {\bfseries} 
  {.} 
  {.5em} 
  {} 
\theoremstyle{mainstyle} 

\newtheoremstyle{exampstyle}
  {4pt} 
  {4pt} 
  {} 
  {} 
  {\bfseries} 
  {.} 
  {.5em} 
  {} 
\theoremstyle{exampstyle} 

\theoremstyle{mainstyle}
\newtheorem{definition}{Definition}

\newtheorem{proposition}{Proposition} 
\newtheorem{corollary}{Corollary} 

\theoremstyle{exampstyle}
\newtheorem{example}{Example}
\newtheorem*{example*}{Example}

\renewcommand{\u}{u}
\newcommand{\y}{y}
\newcommand{\Y}{Y}
\newcommand{\x}{\bm{x}}
\newcommand{\X}{\bm{X}}

\newcommand{\z}{\bm{z}}
\newcommand{\Z}{\bm{Z}}

\DeclareMathOperator*{\E}{\mathbb{E}}
\DeclareMathOperator*{\p}{\textit{P}}

\DeclareMathOperator*{\argmin}{arg\,min}

\newcommand{\U}{U}

\begin{document}

\title{Learning from Discriminatory Training Data}


\author{Przemyslaw Grabowicz}
\authornote{Authors contributed equally to this research.}
\affiliation{%
  \institution{University of Massachusetts Amherst}
  \city{Amherst}
  \state{MA}
  \country{USA}}
\email{grabowicz@cs.umass.edu}

\author{Nicholas Perello}
\authornotemark[1]
\affiliation{%
  \institution{University of Massachusetts Amherst}
  \city{Amherst}
  \state{MA}
  \country{USA}}
\email{nperello@umass.edu}

\author{Kenta Takatsu}
\affiliation{%
  \institution{Carnegie Mellon University}
   \city{Pittsburgh}
   \state{PA}
  \country{USA}}
\email{ktakatsu@andrew.cmu.edu}


\renewcommand{\shortauthors}{Grabowicz, et al.}

\begin{abstract}

Supervised learning systems are trained using historical data and, if the data was tainted by discrimination, they may unintentionally learn to discriminate against protected groups. 
We propose that fair learning methods, despite training on potentially discriminatory datasets, shall perform well on fair test datasets. Such dataset shifts crystallize application scenarios for specific fair learning methods. For instance, the removal of direct discrimination can be represented as a particular dataset shift problem. For this scenario, we propose a learning method that provably minimizes model error on fair datasets, while blindly training on datasets poisoned with direct additive discrimination.
The method is compatible with existing legal systems and provides a solution to the widely discussed issue of protected groups' intersectionality by striking a balance between the protected groups.
Technically, the method applies probabilistic interventions, has causal and counterfactual formulations, and is computationally lightweight --- it can be used with any supervised learning model to prevent direct and indirect discrimination via proxies while maximizing model accuracy for business necessity.
\end{abstract}

\ccsdesc[500]{Computing methodologies~Machine learning algorithms}
\ccsdesc[100]{Applied computing~Law, social and behavioral sciences}
\ccsdesc[500]{Computing methodologies~Supervised learning}

\keywords{supervised learning, algorithmic fairness, discrimination, dataset shift, concept shift, law, explainability, intersectionality, evaluation}

\maketitle

\section{Introduction}
With the growth of algorithmic decision-making systems in highly consequential domains such as finance and criminal justice, lawmakers have refocused their broader equity agendas to now include assurances that such algorithms do not discriminate~\cite{ntia2023policy}. That is, algorithmic decision-making systems should not treat someone unfavorably because of their membership to a particular group, characterized by a \textit{protected attribute} such as race or gender. Therefore, new guidelines and orders that aim to prevent algorithmic discrimination have been increasingly proposed in recent years, e.g., the U.S. blueprint for an ``A.I. Bill of Rights'' in 2022 \cite{aibill}. 
These proposals are typically based on legal~\cite{titlevii, fairhousing} and social science~\cite{Ture1968Black,Altman2016Discrimination,LippertRasmussen2012Badness} contexts, where the key basis for identifying algorithmic discrimination is whether there is a disparate treatment or unjustified disparate impact on the members of some protected group. 
To prevent disparate treatment, the law often forbids the use of certain protected attributes, $Z$, such as race or gender, in decision-making, e.g., in hiring~\cite{titlevii}. 
Thus, these decisions, $Y$, should be based on a set of relevant attributes, $\X$, and should not depend on the protected attribute, $Z$, i.e., $\p(y|\x,z) = \p(y|\x,z')$ for any $z, z'$, ensuring that there is no \textit{disparate treatment}.\footnote{Throughout the manuscript we use a shorthand notation for probability: $\p(y|\x,z) \equiv \p(Y=y|\X=\x,Z=z)$, where $\X,Y,Z$ are random variables, $\x,y,z$ are their instances, and $\p$ is a probability distribution or density.}
We refer to this kind of discrimination as \textit{direct discrimination} (or lack thereof), because of the direct use of the protected attribute~$Z$.

\begin{figure}[t!]
	\centering
	\includegraphics[width=0.99\columnwidth]{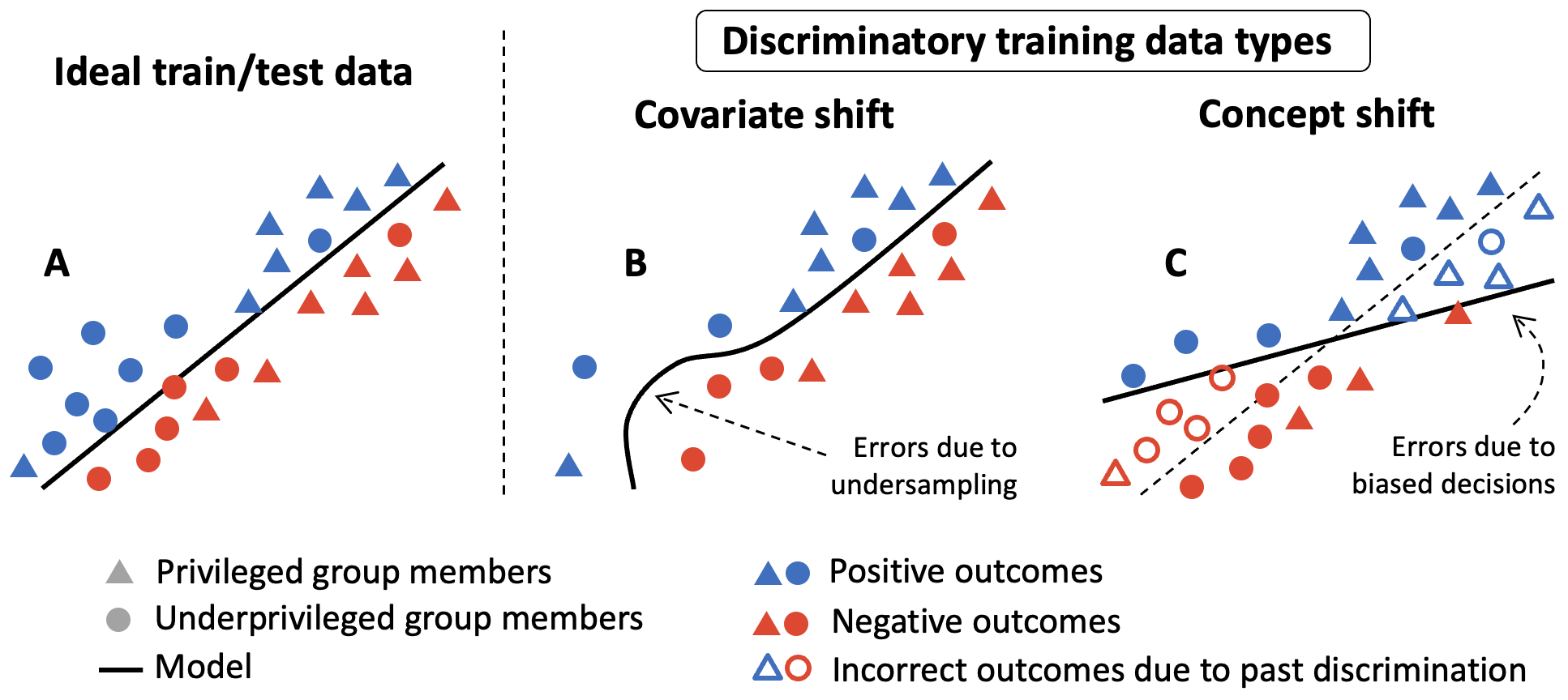}
	\caption{Training data can be tainted in two ways: individuals belonging to underprivileged groups may be undersampled and, hence, models trained on this data may make larger errors for these groups (B), some of the labels in the training data may be incorrect due to historic discrimination and, hence, models trained on this data may be biased against the underprivileged groups (C). These two dataset issues represent a covariate shift and concept shift, respectively. This paper addresses discriminatory concept shifts.}
\vspace{-0.1cm}
	\label{fig:dataset-shifts}
\end{figure}


Despite the introduction of laws prohibiting direct discrimination in the 20th century, such protections were sometimes circumvented by the use of attributes correlated with the protected attribute as proxies. 
One example of this is the practice of ``redlining'' done by U.S. financial institutions. That is, these institutions systematically denied loans and services to customers residing in neighborhoods with populations largely comprised of racial and ethnic minorities~\cite{Zenou2000Racial,Hernandez2009Redlining}.
In order to prevent such \textit{inducement of discrimination} via proxy attributes, legal systems have established that the probability of a positive decision should be the same among individuals belonging to different protected groups~\cite{LippertRasmussen2012Badness,Altman2016Discrimination,titlevii, fairhousing}, i.e., $\p(y|z_1) = \p(y|z_2)$. 
Such protections are also legally necessary for decision-making systems \cite{eo14091}, especially since data-rich machine learning systems can often find accurate surrogates for protected attributes when a large enough set of legitimate-looking variables is available, resulting in discrimination via association~\cite{Wachter2019Affinity}. 
However, these laws often have provisions allowing for such \textit{disparate impact} across groups if there is a ``justified reason'' or ``business necessity clause''~\cite{titlevii}.
For instance, in the 1970s it was found that females were less likely to be admitted than males in graduate admissions to University of California Berkeley~\cite{bickel1975sex}. However, females applied to departments with lower admission rates than males and the overall admissions process was judged legal. 
The provisions allowing for \textit{disparate impact} conflict with the statistical notions of fairness, the fairness definitions most common in algorithmic fairness literature \cite{lipton2019troubling}. These notions typically call for parity of a statistical measure, e.g., impact parity: $\p(y|z_1) = \p(y|z_2)$~\cite{fairmlbook}, which prevents the usage of attributes related to the protected-attribute.
To address the challenge of handling business necessity and proxy attributes, and to develop a method that is transparent and communicable to lawmakers and courtroom officials, our prior work employed explainability measures to remove direct discrimination without the inducement of discrimination~\cite{marrying}. Our prior work, however, did not discuss the real-world setting of multiple protected attributes, did not specify the training dataset issues, and was not optimally accurate --- we address these gaps in this study.

In legal texts, the prevention of discrimination spans across many groups defined over multiple protected attributes, e.g, race, gender, and religion \cite{aibill,titlevii,fairhousing}. Despite this, there rarely exists any legal mechanisms accounting for discrimination based on the intersection of the protected attributes an individual may have --- a concept known as ``intersectionality'' which has been famously spotlighted by social experts in recent decades~\cite{legalintersectionality}. The need for such mechanisms can be seen in criminal justice settings such as COMPAS~\cite{larson2016how}, where it is well documented that certain intersections of age, race, and sex experience more discriminatory outcomes than others, e.g, young Black males~\cite{sentenctingproject}.
With the lack of legal support on preventing discrimination on these intersections, it is unsurprising that many fair learning methods do not operate in such settings and even fewer report results in them~\cite{facctintersectionality}. In this work, we address this setting. Doing so is crucial for algorithmic fairness, as prior studies have shown that learning methods can be fair with respect to protected attributes separately, such as race and sex, while being discriminatory to intersections of attributes, e.g., Black females or Black males~\cite{gerryfair}. 


Another crucial challenge is how to clarify application scenarios of algorithmic fairness methods. With this clarification, policymakers could utilize the information about such scenarios to shape future legislature regulating consequential algorithmic decision-making~\cite{grabowicz2023ai}. Therefore, we propose to distinguish between various data issues and tie them with the methods that address these issues. 
This task has received much less research attention than the fair learning methods themselves. Unfortunately, the research community that studies the data issues for supervised learning, so-called dataset shifts~\cite{Widmer1996Learning, MorenoTorres2012unifying, lu2019learning}, is largely disconnected from the algorithmic fairness community~\cite{fairmlbook}.
In supervised learning, models are trained to perform well on training data and are evaluated on test data, where both are typically created by splitting a dataset into two subsets. In contrast, dataset shifts refer to data issues where there are systematic differences between train and test datasets. 
To our knowledge, we are the first to note that different algorithmic fairness problems can be formalized as different kinds of dataset shifts.
Firstly, if one of the protected groups is underrepresented in the training set, this commonly results in larger model errors for underprivileged group (Figure~\ref{fig:dataset-shifts}B)~\cite{halpern2018gender}. This problem can be formalized as a covariate shift, i.e., $P_{\text{train}}(Z) \neq P_{\text{test}}(Z)$, and it can be solved via sample reweighing or subsampling of the majority group~\cite{sagawa2020investigation}. Secondly, if the training dataset includes examples of discriminating decisions (Figure~\ref{fig:dataset-shifts}C), then we posit that the model should be evaluated on a non-discriminatory test dataset (Figure~\ref{fig:dataset-shifts}A). Formally, this is a concept shift problem, i.e., $P_{\text{train}}(Y|\X,Z) \neq P_{\text{test}}(Y|\X,Z) $, that we address in this work.

\textbf{Problem summary.}
%
Consider decisions $\Y$ that are outcomes of a process acting on non-protected variables $\X$ and protected variables $\Z$, where $\x\in \mathcal{X}$, $\z\in \mathcal{Z}$, $\y\in \mathcal{Y}$, i.e., the variables can take values from any set, e.g., binary or real.
Protected and non-protected features are indexed, e.g., $X_i$ corresponds to the $i$'th feature (component).
We are interested in training a model on available dataset $D_\text{train}$ sampled from $P_\text{train}(\X,\Z,Y)$. 
This model can represent any decision-making process, e.g., 
assigning a credit score for a customer, given their financial record $\x$ and their ethnicity and gender $\z$.
%
The goal of a standard supervised learning algorithm is to obtain a function $\hat{\y}: \mathcal{X} \to \mathcal{Y}$ that optimizes a given objective, e.g., the expected loss, $\E_{D_\text{train}}[\ell( \Y, \hat{\y}(\X) )]$, where the expectation is over the samples in $D_\text{train}$ and $\ell$ is a loss function, e.g., quadratic loss, $\ell(\y,\hat{y})=(\y-\hat{y})^2$.

However, if the training dataset is tainted by discrimination, then a data science practitioner may desire, and, in principle, be obliged by law to apply an algorithm that does not perpetuate this discrimination.
For clarity, we distinguish between discriminatory decisions $T\in \mathcal{Y}$ that are causally and unfairly influenced by~$Z$ (Figure~\ref{fig:dataset-shifts}C) and non-discriminatory $U\in \mathcal{Y}$ that are are not unfairly influenced by $Z$ (Figure~\ref{fig:dataset-shifts}A).
These two kinds of decisions may co-exist in the same context, e.g., a company's hiring team can include both discriminating and non-discriminating members who determine hires in parallel following nearly the same decision-making process.
Unfortunately, the practitioner may have no information whether the training dataset was tainted by discrimination, $D_\text{train}=\tilde{D}=\{(\x^i,\z^i,t^i)\}$, where $i\in \{1,...,n\}$ is a sample index, or was \textit{not}, $D_\text{train}=D=\{(\x^i,\z^i,\u^i)\}$, nor how it was tainted, so supervised algorithms that aim to prevent discrimination operate in a blind setting. 
The problem that we aim to address is to provide a learning algorithm that in such a blind setting yields models that are as close to non-discriminatory data as possible.



\begin{figure}[t!]
	\centering
	\includegraphics[width=0.6\columnwidth]{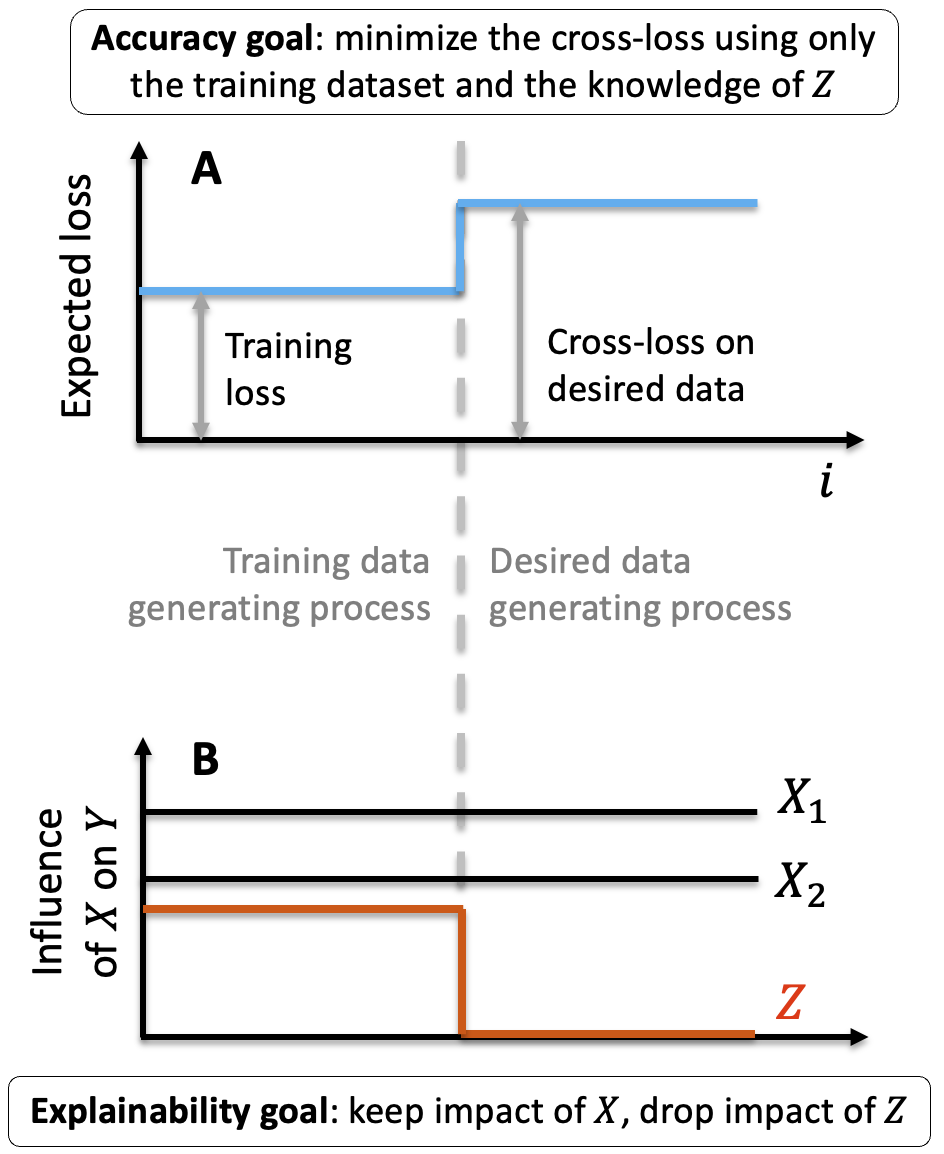}
    \caption{Illustration of the two related goals for fair algorithmic learning, grounded in dataset shifts (top) and explainability literature (bottom). This work focuses on the former, while our prior work focused on the latter.}
    \label{fig:goals}
\end{figure}

\textbf{Contributions.}
%
To address this problem, independently of the given training data type,
we propose that the \textit{objective} of fair supervised learning methods is to minimize the expected \textit{cross-loss}, $\E_{D_\text{test}}[\ell( U, \hat{\y}(\X) )]$, on the non-discriminatory test dataset $D_\text{test}$ drawn from $P_\text{test}(\X,\Z,U)$, while training on a potentially discriminatory data $D_\text{train}$ (\S\ref{sec:defs}), as in Figure \ref{fig:goals}A.
Achieving that objective may sound infeasible, given lack of any assumptions about the concept shift, i.e., we are in the blind setting, but the information that the attribute $\Z$ should not directly influence the model outcomes $\hat{Y}$ is the reason why this problem is solvable.
We show that a learning algorithm averaging probabilistic interventions on the protected attribute optimizes cross-loss under additive directly discriminatory dataset shifts (\S\ref{sec:oip}).
Such interventions previously were applied to compute explainability measures~\cite{Datta2016Algorithmic, Janzing2019Feature}, and were used in the context of discrimination prevention only recently by our work \cite{marrying}. In that study, we proposed that the goal of a fair learning algorithm is to nullify the influence of the protected attribute, while preserving the influence of remaining attributes (explainability goal in Figure \ref{fig:goals}), which is achieved by \textit{marginal interventional mixtures}. In this work, we introduce a novel ``accuracy'' goal of cross-loss minimization, which is achieved by \textit{optimal interventional mixtures}, and show that the two methods are equivalent in certain conditions.
We evaluate and compare the optimal interventional mixture with the state-of-the-art algorithms addressing discrimination (\S\ref{sec:rw}) on synthetic datasets simulating direct discrimination and proxy variables (\S\ref{sec:synthetic}), and on real-world datasets~(\S\ref{sec:realdata}), including those with multiple protected attributes, finding that the \textit{optimal interventional mixture} leverages parity measures and accuracy, and can accurately recover the unbiased ground truth. Our method is included in the publicly released \texttt{FaX-AI} Python library (\url{https://github.com/social-info-lab/FaX-AI}).

\section{Related Works}
\label{sec:rel}

\textbf{Causal notions of fairness}.
One can define direct and indirect discrimination as direct and indirect causal influence of $Z$ on~$Y$, respectively~\cite{Zhang2017causal,Zhang2018Fairness, Marx2019Disentangling}. 
While this notion of direct discrimination is consistent with the concept of disparate treatment in legal systems, the corresponding indirect discrimination is not, since the business necessity clause allows the use of an attribute that depends on the protected feature (causally or otherwise) if only the attribute is judged relevant to the decisions made, e.g., as in the seminal court case of Ricci v.~DeStefano~\cite{ricci}.
This issue is addressed by \textit{path-specific} notions of causal fairness~\cite{Nabi2019Learning, Chiappa2019Path,Wu2019PC}.
However, if there is no limit on the influence that can pass through fair paths, then the path can be used for inducing discrimination, as in the aforementioned case of \textit{redlining}. 
Hence, causal accounts of discrimination~\cite{Kilbertus2017Avoiding, Kusner2017Counterfactual, Zhang2018Fairness, Salimi2019Capuchin, Nabi2019Learning, Chiappa2019Path,Wu2019PC} do not capture induced discrimination, which is common in machine learning and is the focus of this work. 
To address this issue, our recent work defines induced discrimination as a change in the causal influence of non-protected features associated with the protected attributes and proposes a marginal interventional mixture to inhibit direct and induced discrimination \cite{marrying}. However, that work does not discuss multiple protect attributes and it does not consider discriminatory concept shifts.

\textbf{Dataset shifts}.
There is a growing interest in the machine learning community in dataset shifts, since they are surprisingly common in reality and often negatively impact the performance of supervised models on deployment~\cite{koh2020wilds, sagawa2020investigation}. The most common dataset shift is a covariate shift, where the distribution of features or decisions changes between the training and test datasets, i.e., $P_{\text{train}}(\x,z) \neq P_{\text{train}}(\x,z) $, or $P_{\text{train}}(\y) \neq P_{\text{test}}(\y) $, respectively~\cite{MorenoTorres2012unifying}.
In the context of fair machine learning, outcome perturbations were first proposed as random swaps of labels in binary classification, i.e., $\y \sim \p(\y|\u)$, where $\y$ is a perturbed version of $\u$~\cite{Fish2016confidence}. 
That study, however, assumed no access to the protected attribute, so the random swaps correspond to adding i.i.d. noise in the output variable. 
Here, we propose to use a different type of dataset shift, known as concept shift, i.e., $P_{\text{train}}(y|\x,z) \neq P_{\text{test}}(y|\x,z) $, to simulate discriminatory perturbations of data and evaluate the resilience of learning methods to such perturbations.

\section{Problem formulation}
\label{sec:defs}
Before we formalize the problem of discrimination prevention based on dataset shifts, we must first define discrimination in the context of decision making. While many other studies focus on statistical notions of fairness \cite{Donini2018Empirical, Woodworth2017Learning, Hardt2016Equality, Pleiss2017Fairness,Zafar2017Fairness, fairmlbook}, our dataset shift-based notions are drawn from abstractions of legal concepts and causal influence notions.

\subsection{Fairness and discrimination}
Our prior work defines unfair influence and fair relationship between protected attributes $\Z$ and decisions $\Y$ by tying them to legal texts and instruments~\cite{marrying}. 
\begin{definition}
\textbf{Unfair influence} is an influence of protected feature(s) $\Z$ on specified type of decisions $\Y$ that is judged illegal via some legal instrument, e.g., Title VII of the U.S. Civil Rights Act~\cite{titlevii}. 
\end{definition}
\begin{definition}
\textbf{Fair relationship} of protected feature(s) $\Z$ with non-protected feature(s) $\X$ is a relationship that is judged legal when making decisions $\Y$, e.g., due to the U.S. business necessity clause. 
\end{definition}
In real-world contexts, many models can generate decisions $\Y$ without directly using the protected attribute $\Z$, while using non-protected features $\X$ which may be associated with the protected attribute. Even though these features may be related to the protected attribute, they may be legally admissible for use in the decision-making if they are not \textit{unfairly influenced} by the protected feature(s), i.e., they are relevant to the decisions and fulfil a business purpose recognized by legal agencies.
For instance, in the case of Ricci v. DeStefano~\cite{ricci}, the U.S. Supreme Court ruled that the feature in question, a promotion exam, did not violate business necessity despite its association with race. Thus according to the court, there was a \textit{fair relationship} between the exam and race.

With these definitions of unfair influence and fair relationship, discrimination can be defined through measures of causal influence. Formal frameworks for causal models include classic potential outcomes (PO) and structural causal models (SCM)~\cite{pearl2009causality}. In this notation, the potential outcome for variable $\Y$ after intervention $do(\X=\x,\Z=\z)$ is written as $Y_{\x,\z}$, which is the outcome we would have observed had the variables $\X$ and $\Z$ been set to the values $\x$ and $\z$ via an intervention. 
It is assumed that there are direct causal links from $\X$ and $\Z$ to $\Y$, that all variables are observed, and there are no assumptions about the relations between $\X$ and $\Z$ and their components.
These assumptions hold at the very least for a model $\hat{Y}$ of $\Y$ that uses $\X$ and $\Z$ as features. This foundational point enables explainability measures, e.g., various feature influence definitions~\cite{Janzing2019Feature}. Hence, in our prior work we argue that if the intentions and reasoning behind the development process of the model $\hat{Y}$ \textit{was} legally admissible, e.g., proxies were not used as a replacement for the protected attribute, then despite the unknowingly incorrect epistemic state represented by the model, e.g., partially incorrect causal representation, legal systems may acquit model developers of discrimination \cite{marrying}.
Under these assumptions, the causal \textit{controlled direct effect} (CDE) on $\Y$ of changing the value of $\Z$ from a reference value $\z$ to $\z'$ given that $\X$ is set to $\x$~\cite{pearl2009causality} is
\begin{align}
    \text{CDE}_{\Y}(\z',\z | \x) = \E[ \Y_{\x,\z'} - \Y_{\x,\z}].
\end{align}
By tying the causal concept of controlled direct effect to the notions of \textit{fair influence} and \textit{unfair relationship}, we define three concepts of discrimination -- direct, indirect, and induced \cite{marrying}.

\begin{definition}
\textbf{Direct discrimination} is an \textit{unfair influence} of protected attribute(s), $\Z$, on the decisions \mbox{$\Y$, i.e., $\exists_{z,z'} \exists_{\x} \text{CDE}_{\Y}(\z,\z' | \x) \neq 0$}.
\end{definition}

\begin{definition}
\textbf{Indirect discrimination} is an influence on the decisions $\Y$ of feature(s)~$\X$ whose \textit{relationship} with $\Z$ is not \textit{fair}, i.e., $\exists_{x,x'} \exists_{\z} \text{CDE}_{\Y}(\x,\x' | \z) \neq 0$.
\end{definition}

\begin{definition}
\textbf{Discrimination induced} via~$X_i$ is a transformation of the process generating decisions $U$ not affected by direct and indirect discrimination into a new process generating $\Y$ that modifies the influence on the decisions of certain $X_i$ depending on $\Z$ between the processes $U$ and $Y$, i.e., 
$\exists_{\z} \exists_{\x,\x'} 
\text{CDE}_{U}(\x,\x' | \z) \neq \text{CDE}_{\Y}(\x,\x' | \z)$ given that 
$P(\x|\z) \neq 
P(\x)$ or $P(\x'|\z) \neq 
P(\x')$.
\end{definition}

To remove direct discrimination, one can construct a model~$\hat{Y}$ that does not use $\Z$.
However, this may induce discrimination indirectly via the attributes $X_i$ associated with the protected attributes $\Z$, even if there is no causal link from $\Z$ to $X_i$. Methods inhibiting discrimination should do so without inducing discrimination.

\begin{example}
Consider a hypothetical linear model of loan interest rate, $\Y$. Using similar models, prior works suggest that interest rates differ by race, $Z$~\cite{Turner1999Mortgage, Bartlett2019Consumer-Lending}. Some loan-granting clerks may produce non-discriminatory decisions, $u=\beta_0-x_1$, while other clerks may discriminate directly, $y_\text{dir}=\beta_0-x_1+z$, where $\beta_0$ is a fixed base interest rate, $x_1$ is a relative salary of a loan applicant, while $z$ encodes race and takes some negative (positive) value for White (non-White) applicants. 
If the protected attribute is not available, e.g., loan applications are submitted online, then a discriminating clerk may induce discrimination in the interest rate, by using a proxy for race, $y_\text{ind}=\beta_0-x_1+x_2$, where $x_2$ is the proxy, e.g., an encoding of the zip code (as in the redlining) or the first name (as in the seminal work of~\citeauthor{Bertrand2003Are}) of the applicant. 
\end{example}

\subsection{Discriminatory concept shifts}

Distinct from our prior work, we introduce an additional goal in discrimination prevention from the perspective of dataset shifts. That is, we propose to use discriminatory perturbations dependent on the protected attribute (or all possible intersections of multiple protected attributes) to simulate a concept shift, i.e., $P_{\text{train}}(y|\x,z) \neq P_{\text{test}}(y|\x,z) $, and to evaluate the cross-loss of learning methods w.r.t. to such concept shifts~\cite{MorenoTorres2012unifying} (accuracy goal in Figure \ref{fig:goals}).
These concept shifts reflect bias in a historical data-generating process, rather than a sampling bias which typically is associated with covariate shifts.
\begin{definition}
\label{def:conceptshift}
\textbf{Discriminatory concept shift} is a transformation of the process generating $U$ that is not affected by direct, indirect, and induced discrimination into a new process generating $\Y$ that is affected by discrimination.
\end{definition}

\begin{example}
    We continue the prior example. The transformation from $u=\beta_0-x_1$ to $y_\text{dir}=\beta_0-x_1+z$ via a directly discriminatory additive perturbation of $z$ (race) is a discriminatory concept shift. This gives two datasets, $\tilde{D} = \{( x_1^i, x_2^i, z^i, y_\text{dir}^i )\}$ for training and $D = \{( x_1^i, x_2^i, z^i, u^i )\}$ for testing.
\end{example}

We do not assume that the perfectly fair decision-making process, illustrated in Figure \ref{fig:dataset-shifts}A, exists already in all real-world contexts. In stark contrast, we posit that its knowledge should not be required to prevent discrimination in supervised learning. The above constructs enable us to formalize the goal for fair learning methods on the grounds of dataset shifts and specify the idealized real-world scenarios that the methods achieving this goal address. 
Next, we define the cross-loss of a supervised learning algorithm to discriminatory concept shifts, which measure how well an algorithm trained on \textit{potentially} discriminatory training dataset, i.e.,~$D_\text{train} = \tilde{D}$ or $D_\text{train} = D$, performs when it is evaluated on a non-discriminatory~$D_\text{test}=D$.

\begin{definition}
\textbf{Cross-loss.}
The solution of supervised learning algorithm $a$, $\hat{y}_a(\x|D_\text{train})$, is a model obtained by training on the potentially discriminatory dataset $D_\text{train}$.
The empirical \textit{cross-loss} function is an expected loss of this model w.r.t. the non-discriminatory data $D$, 
$
\E_D\left[ \ell\left( U, \hat{y}_a( \X | D_\text{train} )\right) \right].
$
\end{definition}

The cross-loss measures how well the model learned by an algorithm training on the discriminatory data predicts the fair data, i.e., how well it performs under a discriminatory concept shift. 

\begin{example}
\label{ex:cross}
We continue the prior example.
For simplicity, assume that all variables have zero mean, 
no correlation between $X_1$ and $Z$, and a positive correlation $r>0$ between $X_2$ and $Z$. Let the training dataset be $\tilde{D} = \{( x_1, x_2, z, y_\text{dir} )\}$. If we applied standard supervised learning under the quadratic loss, then asymptotically with the number of samples we would learn the model $\hat{y}_1 = \beta_0 - x_1 + z$, which is directly discriminatory and results in high cross-loss $\E_D\left[ \ell\left( U,\hat{y}_1(\X|\tilde{D})\right)\right] = \E_Z Z^2$. If we dropped the protected attribute, $Z$, before regressing $Y_\text{dir}$ on the attributes $X_1$ and $X_2$, then we would learn the model $\hat{y}_2 = \beta_0 - x_1 + r x_2$, which also yields a sub-optimal cross-loss, $\E_D\left[ \ell\left( U,\hat{y}_2(\X|\tilde{D})\right) \right] = r^2 \E_{X_2} X_2^2$, that increases with $r$ due to the growing discrimination induced via $X_2$. 
\end{example}

%
%

\section{Optimal interventional mixture}
\label{sec:oip}

Next, we introduce a supervised learning method based on probabilistic interventions that aims to prevent direct discrimination in $\Y$ without inducing any discrimination. We prove that it minimizes cross-loss, up to a constant, under the assumption of the concept shift coming from additive directly discriminatory perturbations~(\S\ref{sec:oim}).
In addition, if $\Y$ is impacted by \textit{indirect discrimination}, i.e., $\Z$ unfairly influences $\X$, we can address it as \textit{direct discrimination} in $\X$. To prevent \textit{indirect discrimination} one can apply our method in a nested way~(\S\ref{sec:indirect-removal}) that resembles the path-specific counterfactual fairness~\cite{Chiappa2019Path}.

\subsection{Removal of direct discrimination}
\label{sec:oim}

%
The proposed method is a post-processing approach and has two optimisation steps. 
In the first step, we train the model $\hat{y}(\x,z)$  using all features, both protected $Z$ and relevant $\X$, without any consideration of fairness, by minimizing the corresponding expected loss $\E_{D_\text{train}}[\ell(Y,\hat{y}(\X))]$. Most importantly, the protected attribute is available during the training, so the model does not need to use third variables as surrogates of the protected attribute and avoids inducing discrimination via $\X$ (we provide theoretical and empirical evidence for this statement in Proposition 1 and Section~\ref{sec:resilience-vs-induced}, respectively).
%
In the second step, we eliminate the influence of the protected attribute. This is achieved by intervening probabilistically on the full model trained with all features and mixing the interventions on the protected attribute independent from other variables via a mixing distribution $\pi(Z')$, yielding 
$
\hat{y}_\pi(\x)=\sum_{z'} \hat{y}(\x,z')\pi(z')
$. 
Here, we search for the optimal mixing distribution, $\pi^*(z')$, that minimizes the expected loss, $\E_{D_\text{train}}[\ell(Y,\hat{y}_\pi(\X))]$, while all parameters of the full model $\hat{y}(\x,z)$ are fixed, i.e.,
$
\pi^* = \argmin_{\pi} \E_{D_\text{train}}[\ell( \Y, \hat{\y}_\pi(\X) )].
$
This optimization problem is convex for quadratic and negative log-likelihood loss functions. Thus, the optimal weighting distribution can be found by applying disciplined convex programming with constraints ensuring that $\pi(z')$ is a distribution, i.e., $\sum_{z'} \pi(z') = 1$ and $\pi(z')\geq 0$ for all $z'$~\cite{Diamond2016CVXPY}. 
Once the optimal mixing distribution is known, the \textit{optimal interventional mixture (OIM)} can be computed,
$
\hat{y}^*(\x)=\sum_{z'} \hat{y}(\x,z')\pi^*(z'),
$
which constitutes the solution of the proposed learning algorithm.

Unlike many methods achieving statistical fairness objectives, our method is seamlessly applicable to scenarios with multiple protected attributes or numeric attributes such as age. This is accomplished by mixing the interventions on all combinations of the protected attributes in the second optimization step. 
Next, for discriminatory data transformations that have a simple additive form, i.e., $y = u + h(z)$, we prove that optimal interventional mixture minimizes cross-loss on non-discriminatory data and show that for $\ell^2$ loss the accuracy and explainability goals of fair machine learning (Figure~\ref{fig:goals}) lead to the same solution. 



\begin{proposition}
Let the non-discriminatory data have $u=f(\x)+\nu$ and the data following a discriminatory concept shift have $y = f(\x) + h(\z) + \nu$, where $f$ and $h$ are some functions and $\nu$ is i.i.d. noise independent from $\X$ and $Z$. 
Assume that the same $\ell^p$ loss, either $\ell^1$ or $\ell^2$, is used for model learning and the computation of cross-loss.
If the estimation model is well specified w.r.t. the discriminatory data-generating process and the estimation method is consistent, then the OIM, asymptotically with the number of samples, is $\hat{\y}^*(\x) = f(\x)+C_p$, and it minimizes the expected cross loss $\E_D\left[ \ell\left( U, \hat{y}_a( \X | \tilde{D} )\right) \right]$ up to the constant $C_p$ that depends on the unknown $h(\Z)$.
\end{proposition}

\begin{example}
We continue the loan interest rate example. The full model is $\hat{y}(\x,\z) = \beta_0 - x_1 + \z$. The optimal interventional mixture is $\hat{y}^* = \beta_0 - x_1+\beta_\pi$, where the intercept $\beta_\pi$ is the result of mixing over the optimal $\pi^*(z')$. 
In this case, $\beta_\pi=\E_Z Z=0$ due to the optimization. 
Thus, the algorithm recovers the non-discriminatory ground truth.
\end{example}

The proof follows from the definition of consistent estimator (full proof in Appendix A).
For a particular dataset that does not meet the condition $C_p=0$, one can propose a better model than the OIM by subtracting $C_p$ from model's intercept, which is a sum of $C_p$ and a component of $f(x)$, but $C_p$ depends on the unknown $h(Z)$ and, without knowing $h(Z)$, we do not know what to subtract, so there is no learning strategy that improves the cross-loss.
Furthermore, the case of nonzero $C_p$ is practically irrelevant, because it represents a data perturbation that affects all individuals in the same way, e.g., it introduces across the board more positive outcomes $y$ without changing their dependence on $\x$, i.e., $\E[\Y|\x] = \E[U|\x]+C_p$.
The above proposition is valid for well-specified models. Next, we prove analogue result for universal approximators such as deep learning models.




\begin{corollary}
Let the same assumptions hold as in Proposition 1, but now the estimation model is a universal approximator. Then the OIM is an arbitrarily close approximation of $f(\x) + C_p$, which according to Proposition~1 minimizes the expected loss $\E_D\left[ \ell\left( U, \hat{y}_a( \X | \tilde{D} )\right) \right]$ up to $C_p$.
\end{corollary}


The proof follows from universal approximator theorems and Proposition 1 (see Appendix A).
These guarantees do not universally hold for our prior work, which is the only work that proposes a similar interventional mixtures for inhibiting discrimination \cite{marrying}. Rather than finding an optimal mixture, we previously proposed to utilize the marginal distribution of the protected attribute to build a marginal interventional mixture (MIM), i.e., $\hat{y}_\text{MIM}(\x) = \E_{\Z} [\hat{y}(\x,\Z)]$. 

\begin{proposition}
\label{prop:mim}
Let the same assumptions hold as in Proposition 1. Then the marginal interventional mixture (MIM), asymptotically with the number of samples, is $\hat{y}_\text{MIM}(\x) = \E_{\Z} [\hat{y}(\x,\Z)]=f(\x)+\E[h(Z)+\nu]$, and minimizes the expected cross loss $\E_D\left[ \ell\left( U, \hat{y}_a( \X | \tilde{D} )\right) \right]$ for $\ell^2$ loss up to the constant $\E[h(\Z)+\nu]$.

\end{proposition}

%


\subsection{Removal of indirect discrimination via optimal counterfactual mixture}
\label{sec:indirect-removal}

In real-world scenarios, a non-protected feature, $X_i$, can be unfairly influenced by $\Z$. If decisions $Y$ were influenced by such $X_i$, then $Y$ would be indirectly discriminatory. To prevent this, one can apply a nested multi-stage version of OIM. 
More precisely, say that we have $X_1$, $X_2$, and $\Z$, where $X_1$ is unfairly influenced by $\Z$, and all are used to make decisions $Y$. We first create a model $\hat{Y}$ using $X_1$, $X_2$, and $\Z$. Then, we create a model $\hat{X}_1$, using $X_2$ and $\Z$ and other relevant features that we have access to, and apply the OIM to create a ``fair'' model $\hat{X}_1^*$. Lastly, to create $\hat{Y}^*$, we replace $X_1$ with $\hat{X}_1^*$ in the model $\hat{\Y}$, and apply the OIM. This is a reasonable solution, but in situations where we know the value of a variable for which we apply OIM, such as $X_1$ here, we can do better through counterfactual analysis.

\subsubsection{Counterfactual mixtures. }
\label{sec:counterfactual}

Causality literature posits a causal hierarchy and distinguishes between interventional and counterfactual estimates~\cite{Pearl2016Causal}. The latter differ from former in that they assume that everything stays the same, including any exogenous noise values, when estimating the effect of an intervention.
In contrast, the interventional mixture calculates the value of $\hat{X}_1$ had the causal influence of $\Z$ been removed from it given the values of all \textit{observed} variables, but not the values of exogenous noise. 
Each variable can contain \textit{exogenous} noise, i.e., unobserved intrinsic noise not associated with any other variable. In the situations where we know the value of the variable for which we want to develop a fair model, we can use that value to infer that variable's exogenous noise.
For such situations, we propose an \textit{optimal counterfactual mixture} (OCM), which merges the three canonical counterfactual reasoning steps with the OIM step: (\textit{abduction}) infer exogenous noise for a variable, (\textit{intervention}) apply the OIM to remove the influence of the protected attribute on that variable, and (\textit{counterfactual prediction}) estimate the counterfactual value of the variable given the exogenous noise and intervention.

\subsubsection{Counterfactual mixtures comparison. }

We compare the interventional (OIM) and counterfactual (OCM) versions of our method as well as the related path-specific counterfactual fairness (PSCF) using a  multi-stage linear model introduced in the PSCF paper~\cite{Chiappa2019Path}:
\begin{align}
M &= \theta^m + \theta^m_z Z + \theta^m_c C + \epsilon_m, \label{eq:m_pscf}\\
L &= \theta^l + \theta^l_z Z + \theta^l_c C + \theta^l_m M + \epsilon_l, \\
Y &= \theta^y + \theta^y_z Z + \theta^y_c C + \theta^y_m M + \theta^y_l L + \epsilon_y,
\end{align}
where $C$, $M$, $L$ are components of~$\X$, $\Z$ is the protected attribute, and $\epsilon_c$, $\epsilon_m$, $\epsilon_l$ are exogenous noise variables. The causal influence of $Z$ on decisions $\Y$ and the mediator $M$ is assumed unfair and all other influences are fair. In other words, $Y$ is affected by direct discrimination via $Z$ \textit{and} indirect discrimination via $M$. This means that our method needs to be applied first to $M$ and then to $Y$.

For simplicity, without loss of generality, let us consider a scenario where we have enough samples to have perfect estimates of a well-specified model's parameters, so that the estimated model is $\hat{m}=\theta^m + \theta^m_z z + \theta^m_c c$.
In this scenario, the \textit{abduction} step corresponds to computing $\epsilon_m = m-\hat{m}$; the \textit{intervention} step to applying OIM to $\hat{m}$, yielding $\hat{m}^* = \theta^m + \theta^m_z z^* + \theta^m_c c$; and the \textit{counterfactual prediction} to injecting the abducted noise into the estimated model, $\hat{m}^c = \theta^m + \theta^m_z z^* + \theta^m_c c + \epsilon_m$.
Overall, we refer to these three steps as the single-stage OCM.
Same as the PSCF, the multi-stage OCM corrects the decision through a correction on all the variables that are influenced by the protected attribute along unfair pathways.
Thus, we first apply the OCM to get a non-discriminatory counterfactual $\hat{m}^\text{c}$, then we propagate $\hat{m}^\text{c}$ to its descendants and apply the OCM to yield a fair counterfactual $\hat{l}^\text{c}$, and finally we propagate the two counterfactuals to $\hat{y}$ and apply the OIM (not OCM, since we do not observe $Y$) to get~$\hat{y}^\text{c}$:
\begin{align}
\hat{m}^\text{c} &= \theta^m + \theta^m_z z^* + \theta^m_c c + \epsilon_m= m - \theta^m_z(z-z^*),\\
\hat{l}^\text{c} &= \theta^l + \theta^l_z z + \theta^l_c c + \theta^l_m \hat{m}^\text{c} + \epsilon_l = l - \theta_m^l(m-\hat{m}^\text{c}),\\
\hat{y}^\text{c} &= \theta^y + \theta^y_z z^* + \theta^y_c c + \theta^y_m \hat{m}^\text{c} + \theta^y_l \hat{l}^\text{c},
\end{align}
where $z^*$ is the expected value of Z resulting from the optimal mixing distribution for $Z$.
Conversely, applying solely the OIM to obtain $\hat{m}^*$, $\hat{l}^*$, and $\hat{y}^*$ does not take advantage of estimating the noise terms $\epsilon_m$ and $\epsilon_l$, and results in estimators 
\begin{align}
\hat{m}^\text{*} &= \theta^m + \theta^m_z z^* + \theta^m_c c,\\
\hat{l}^\text{*} &= \theta^l + \theta^l_z z + \theta^l_c c + \theta^l_m \hat{m}^\text{*},\\
\hat{y}^\text{*} &= \theta^y + \theta^y_z z^* + \theta^y_c c + \theta^y_m \hat{m}^\text{*} + \theta^y_l \hat{l}^\text{*}.
\end{align}
When comparing $\hat{y}^*$ and $\hat{y}^\text{c}$ we observe that difference in estimating $\epsilon_m$ unsurprisingly yields the noise terms,
$\hat{y}^\text{c} = \hat{y}^* + \theta^y_m\epsilon_m + \theta^y_l \theta^l_m\epsilon_m$, which results in a larger error w.r.t. $Y$ for the OIM than the OCM,
\begin{align}
 \E(Y-\hat{Y}^*)^2 
 = \E(Y-\hat{Y}^\text{c})^2 + (\theta^y_m\epsilon_m + \theta^y_l \theta^l_m\epsilon_m)^2 \label{eq: mse_ocm}.
\end{align}
%
A comparison with the PSCF reveals that $\hat{y}^\text{c} = \hat{y}^{\text{PSCF}} + \Delta$, where $\Delta = z^*( \theta^y_z + \theta^y_m \theta^m_z + \theta^y_l \theta^l_m \theta^m_z )$. The mean squared error w.r.t. $Y$ is larger for the PSCF than for the OCM by the square of the difference, i.e., $\E(Y-\hat{Y}^{\text{PSCF}})^2 = \E(Y-\hat{Y}^c)^2+\Delta^2$. Overall, the OCM is more accurate than the PSCF, because the PSCF relies on a choice of reference value, $z'$, also known as baseline, which is assumed $z'=0$ in the PSCF paper and above example. However, this choice is arbitrary and it is not clear what the baseline should be for non-binary $\Z$. By contrast, the OCM introduces a distribution $\pi(z')$ and optimizes it for accuracy. In addition, it follows from Proposition 1 and Corollary 1, that the OIM and by extension the OCM, are the most accurate interventional and counterfactual models on the non-discriminatory test datasets (up to the unlearnable constant~$C_p$).

\section{Evaluation method and evaluated methods}
\label{sec:rw}

In the remaining sections, we measure the \textit{resilience} of various learning methods to discriminatory concept shifts that have more complex functional forms than the additive shifts described in the previous section. We begin by introducing the notion of resilience and the evaluated learning methods addressing discrimination.

\subsection{Resilience}

Note that the range of cross-loss values depends on the dataset and loss function. To make comparisons across datasets, we introduce the measure of resilience by normalizing the inverse of cross-loss, so that the resilience is a number between 0 and 1. For a specific pair of datasets $D_\text{train}$ and $D$, the larger the cross-loss, the lower the resilience of the learning algorithm to the concept shift from training data~$D_\text{train}$.


\begin{definition}
\textbf{Resilience.}
The resilience of algorithm~$a$ to a discriminatory concept shift from non-discriminatory data $D$ to potentialy discriminatory $D_\text{train}$ is a ratio of the expected loss of the standard algorithm training on $D$ and the cross-loss of algorithm $a$ training on $D_\text{train}$:
\begin{equation}
\Omega_a = 
\E_D\left[ \ell\left( U, \hat{\u}(\X|D) \right) \right] / 
\E_D\left[ \ell\left( U, \hat{y}_a( \X | D_\text{train} )\right) \right],
\end{equation}
where $\hat{\u}(\x|D)$ is a model of the non-discriminatory ground truth trained on dataset $D$.
\end{definition}

The enumerator of resilience takes into account that $U$ can be intrinsically random and unpredictable.\footnote{If $U$ is not intrinsically unpredictable, then $\E_D\left[ \ell\left( U, \hat{\u}(\X|D) \right) \right]$ can be zero. In such cases, a small value could be added to the enumerator and denominator of resilience, to prevent it from taking the value of zero. This scenario is uncommon in practice.}
The resilience is confined, $0 \leq \Omega \leq 1$. This property is ensured if both learning algorithms yielding the models $\hat{\u}(\x|D)$ and $\hat{y}_a(\x|\tilde{D})$ optimise the same vanilla objective function, e.g., both optimize expected loss, where the algorithm $a$ adds an extra component to address discrimination.
An algorithm that is perfectly resilient to the discriminatory concept shift yields $\Omega=1$, and $\Omega=0$ otherwise. 

\subsection{Evaluated learning methods}
A number of algorithms addressing discrimination have been developed by adding a constraint or a regularization to the objective function~\cite{Pedreshi2008Discrimination, Feldman2014Certifying, Zafar2015Fairness, Zafar2017Fairness, Hardt2016Equality, Zafar2017Parity, Woodworth2017Learning, Pleiss2017Fairness, Donini2018Empirical}.
Most of these algorithms prevent direct discrimination, but it should come as no surprise that some of them do not prevent the induction of discrimination.
For instance, the algorithms that put constraints on the aforementioned disparities in treatment and impact~\cite{Pedreshi2008Discrimination, Feldman2014Certifying, Zafar2015Fairness} induce ``reverse'' discrimination, by affecting the members of advantaged group and the people similar to them in a non-desirable manner when training on a non-discriminatory dataset $D$~\cite{Lipton2018Does}.
As an example, such ``reverse'' discrimination would result in less job opportunities for similarly qualified short-haired women than long-haired women, because short hair is associated with males and there is a historical correlation between hiring and gender~\cite{Lipton2018Does}. 
%
%
Other studies propose interesting statistical notions of fairness, such as equalized opportunity, $\p(\hat{y}|y=1,z=0)=\p(\hat{y}|y=1,z=1)$, equalized odds, $\p(\hat{y}|y,z=0)=\p(\hat{y}|y,z=1)$ \cite{Donini2018Empirical, Woodworth2017Learning, Hardt2016Equality, Pleiss2017Fairness}, or parity mistreatment, $\p(\hat{y}\neq y|z=0)=\p(\hat{y}\neq y|z=1)$~\cite{Zafar2017Fairness}. 
%
However, prior works reveals the impossibility of simultaneously satisfying multiple non-discriminatory objectives, such as equalized opportunity and parity mistreatment~\cite{Chouldechova2017Fair, Kleinberg2017Inherent,Friedler2016impossibility}. 
There is a need to compare them. 

We evaluate several of such methods in the next section. For this evaluation, we select a diverse set of algorithms that aim to prevent discrimination through different objectives: disparate impact \cite{Zafar2015Fairness, Zafar2017Fairnessa}, disparate mistreatment \cite{Zafar2017Fairness}, preferential fairness \cite{Zafar2017Parity}, equalized odds \cite{Hardt2016Equality}, a convex surrogate of equalized odds~\cite{Donini2018Empirical}, game-theoretic envy-freeness~\cite{Zafar2017Parity}, and a causal database repair ~\cite{Salimi2019Capuchin}. 
We also evaluate a scenario where we prevent discrimination over multiple protected attributes. Here, the only fair-learning method we evaluate against is the method introduced in the fairness gerrymandering paper \cite{gerryfair}, as it considers fairness, based on the best subgroup-fair distribution over classifiers, across infinitely many subgroups.
In all cases but one, we use implementations of these algorithms as provided by the authors. We re-implemented one of these methods~\cite{Zafar2015Fairness} so that it works for the case of continuous $Y$.
In Appendix B, we report these methods' parameters we select. 


\section{Evaluation on synthetic data}
\label{sec:synthetic}

%
In the synthetic setting, we generate random non-discriminatory datasets $D$, containing samples of $U$, and perform a concept shift to create datasets $\tilde{D}$, containing samples of~$Y$.
Then, datasets $D_\text{train}=\tilde{D}$ are used for training, datasets $D$ are used for testing, and we measure the resilience and the feature influence of various learning algorithms preventing discrimination, including the OIM. 
Next, we make these measurements as a function of the correlation between the protected and non-protected attributes, which often causes learning algorithms to induce discrimination via association.
%
%
We also study the setting where there is no discriminatory concept shift, $D_\text{train}=D'$ (a dataset drawn from the same distribution as the test dataset~$D$), but there is a feature correlated with the protected attributes that is fair to use, i.e., permitted by law.
The learning algorithms operate in a blind setting, i.e., they have no information whether $D_\text{train}=D'$ or $D_\text{train}=\tilde{D}$.
Other scenarios where we randomize the parameters of our data generating process or have a discriminatory concept shift under a complex non-linear functional form are available in Appendix E and H, respectively, and yield qualitatively the same results for resilience.

\subsection{Resilience captures induced discrimination}
\label{sec:resilience-vs-induced}

\textbf{Data generation.}
Without loss of generality, the data generating process of $U$ can yield $\E[U|\x]=\sigma(f(\x))$, where $f$ is a potentially non-linear function, and $\sigma$ is a function establishing the respective support for $U$.
For instance, for classification problems $\sigma$ can be a logistic or softmax function, while for regression it can be identity. 
Next, we simulate discrimination as a concept shift from $U$ that in general can be represented as
$\E[\Y|\x]=\sigma(g(\x,z))$, where $g$ is some function.
These concept shifts may or may not be discriminatory, depending on how expected outcomes were shifted: i) no discrimination, if $g(\x,z) =  f(\x)$, ii) direct discrimination, if $g(\x,z)$ depends on $z$, iii) induced discrimination, if $g(\x,z) = \tilde{f}(\x) \neq f(\x)+\text{const}$.
We study simple forms of $f(\x)$ and $g(\x,z)$ that are linear combinations of its arguments, i.e., $f(\x) = \bm{\alpha}^\intercal \x$ and $g(\x,z) = \bm{\tilde{\alpha}}^\intercal \x + \beta z$, and $\sigma$ is the logistic function.


\textbf{Results.}
We focus first on a data-generating process that extends the loan-interest Example to binary dependent variables, which are prevalent in real-world decision-making. 
Specifically, $\u \sim \text{Bernoulli}[\E[U|\x]]$ and $\y \sim \text{Bernoulli}[\E[\Y|\x]]$, where $f(\x) = x_1$ and $g(\x,z)=x_1+\beta z$. 
We model this data with logistic regression and measure how the resilience and the expected value of influence of each feature changes with the increasing correlation between $X_1$ and $Z$. We measure influence using SHAP (SHapley Additive exPlanations), a popular explainability measure \cite{shap}. 

\begin{figure}[t]
\centering
\includegraphics[width=0.5\textwidth]{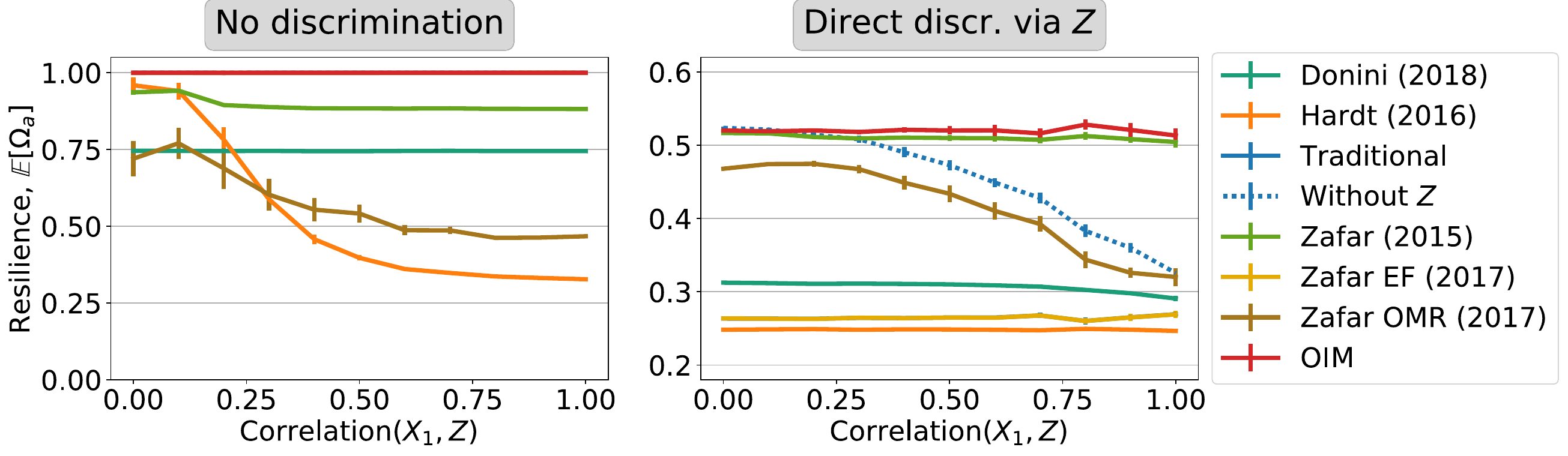}
\caption{
Average resilience to potentially discriminatory concept shifts decreases with the correlation between $X_1$ and $Z$.
The coefficient that scales the discrimination in the training data is $\beta=0$ for the case of no discrimination (left) and $\beta=5$ for direct discrimination (right). 
Each point is an average over 100 random datasets. Error bars show $95\%$ confidence intervals. 
}
\label{fig:corr}
\end{figure}

\begin{figure*}[tb]
\centering
\begin{subfigure}[b]{\textwidth}
  \centering
    \includegraphics[width=.75\linewidth]{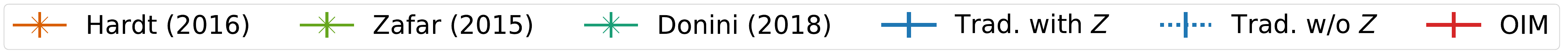}
  \end{subfigure}
  \begin{subfigure}[b]{.49\textwidth}
  \centering
    \includegraphics[width=.49\linewidth]{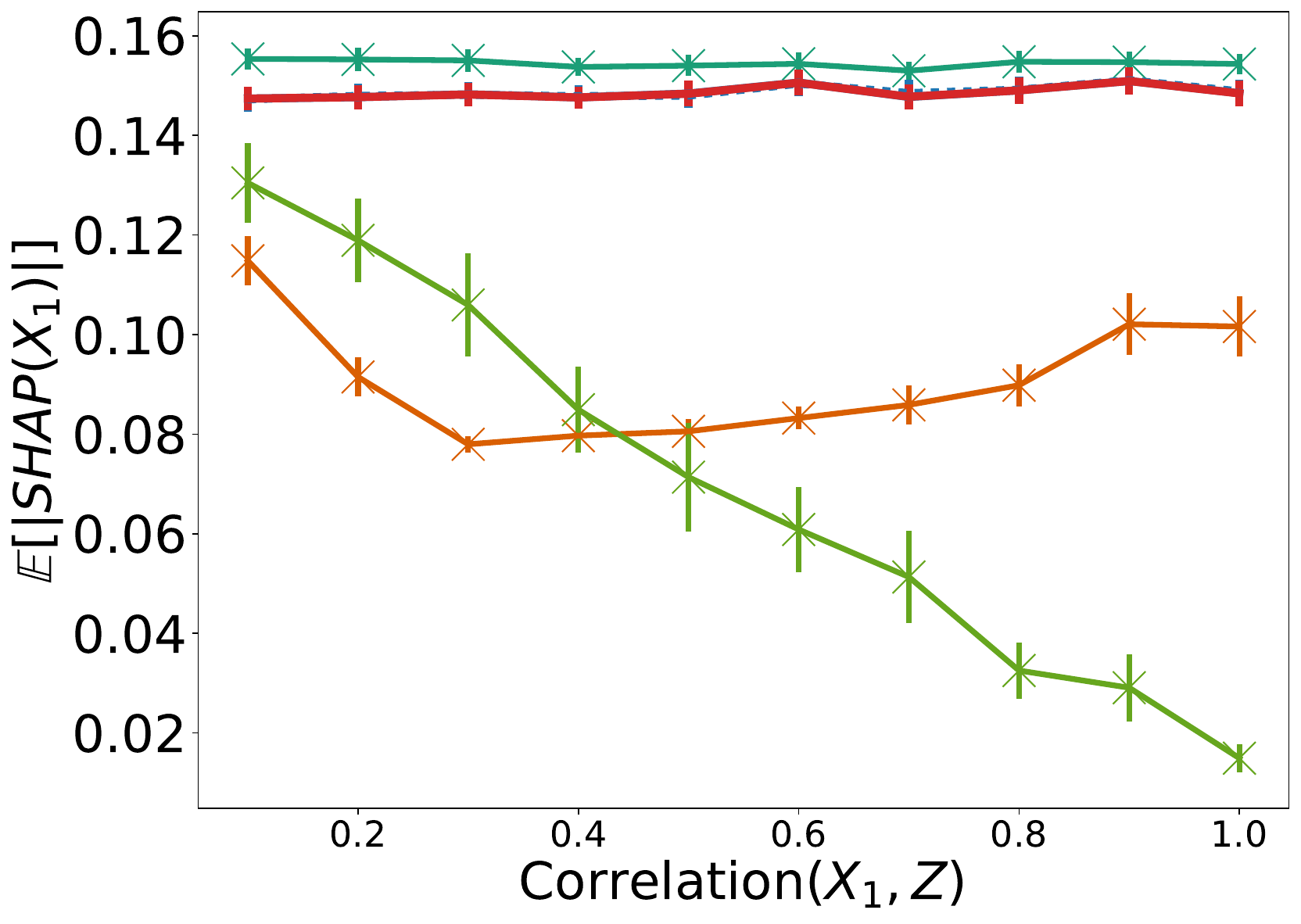}
    \includegraphics[width=.49\linewidth]{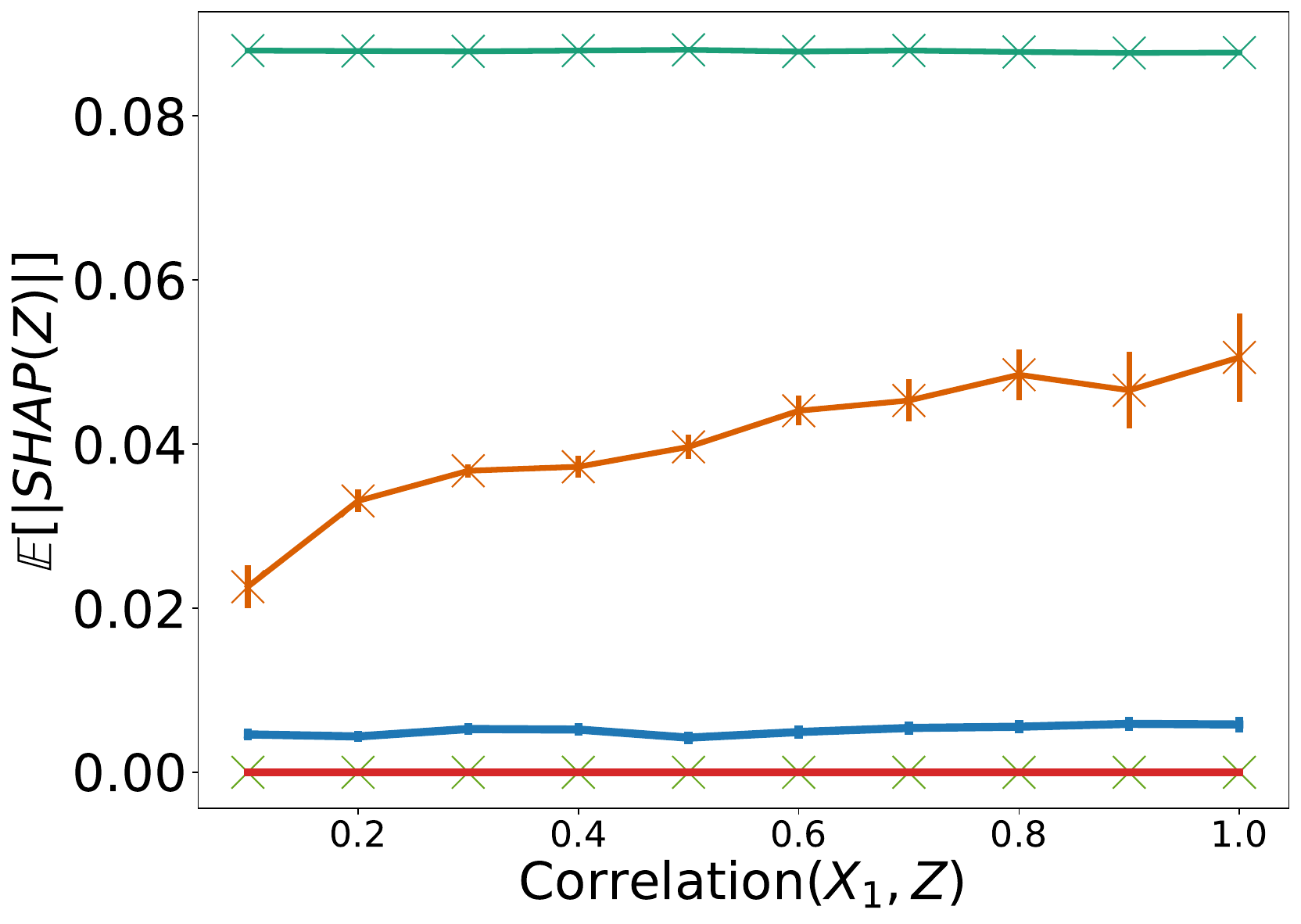}
    \caption{No direct discrimination, $g(\x,z)=x_1+0 * z$ }
  \end{subfigure}
  \begin{subfigure}[b]{.49\textwidth}
  \centering
    \includegraphics[width=.49\linewidth]{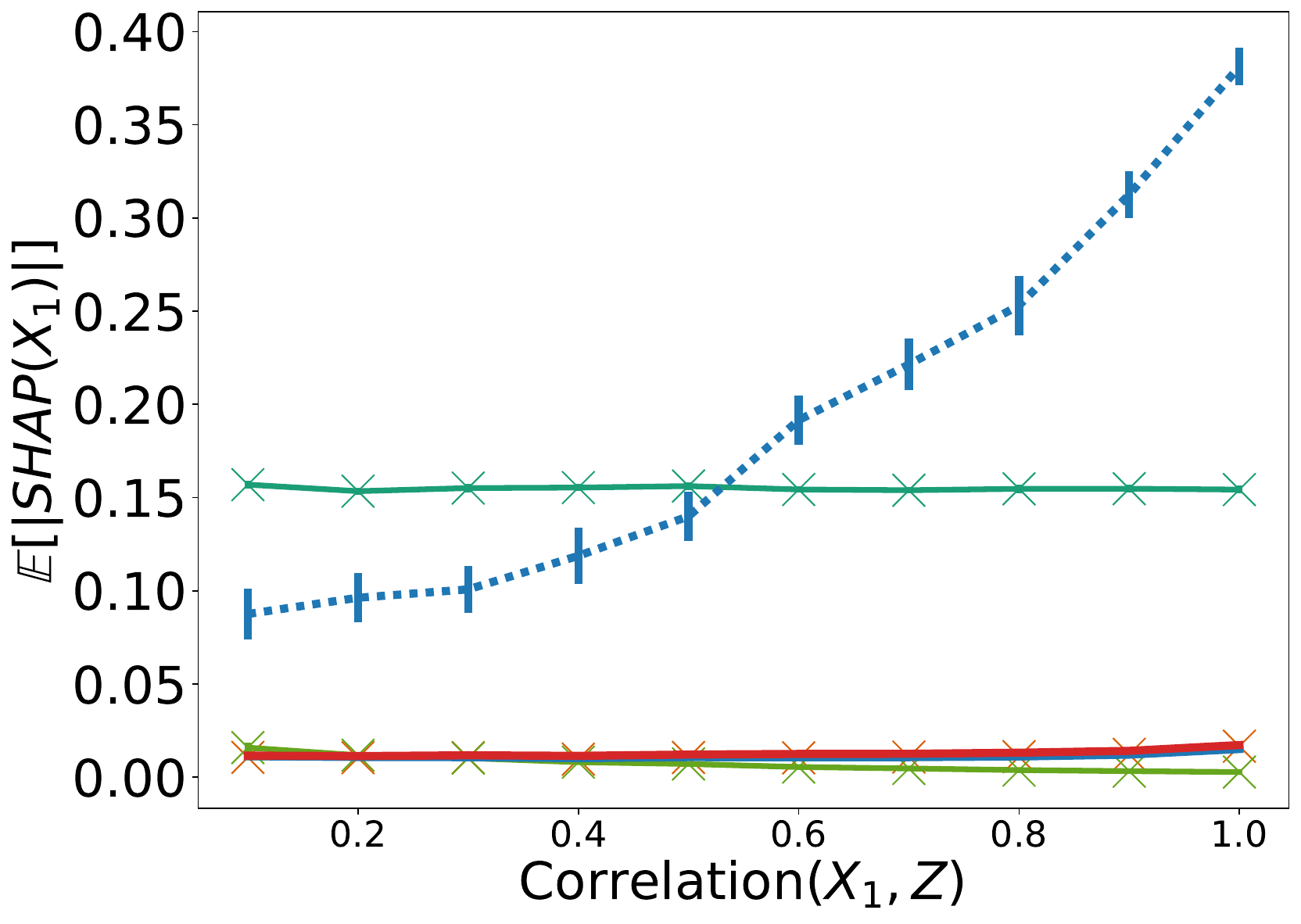}
    \includegraphics[width=.49\linewidth]{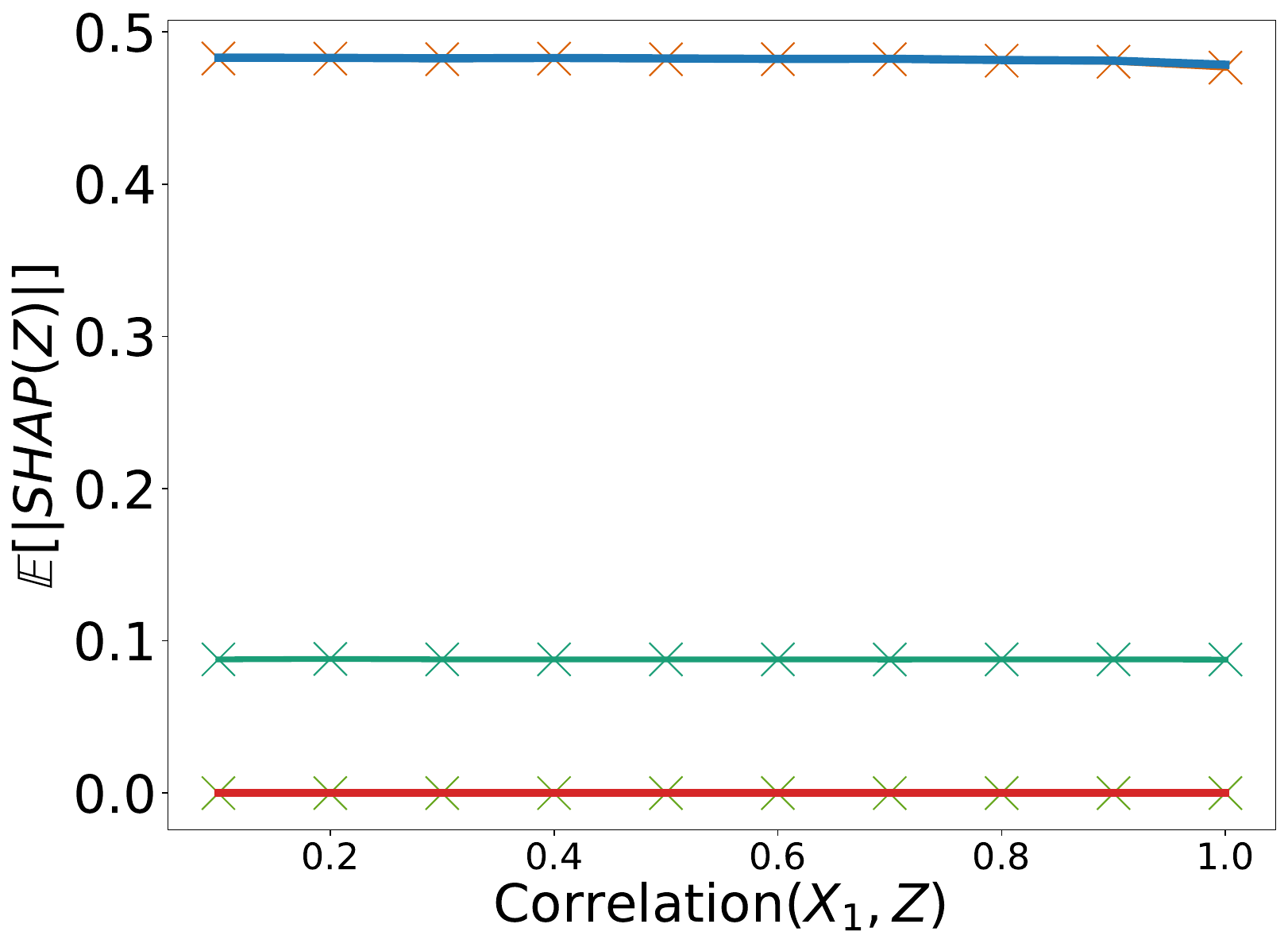}
    \caption{Directly discriminatory, $g(\x,z)=x_1+5 * z$ }
  \end{subfigure}
  
\caption{
Average absolute value of SHAP values for $X_1$ and $Z$ as the correlation between $X_1$ and $Z$ increases. Each point is an average over 100 random datasets. Error bars show $95\%$ confidence intervals. 
}
\label{fig:shap}
\end{figure*}

We study two cases of the training dataset $\tilde{D}$: (i)~without any concept shift (no discrimination, $\beta=0$, left Figures~\ref{fig:corr} \& \ref{fig:shap}) and (ii) with a discriminatory concept shift ($\beta=5$, right Figures~\ref{fig:corr} \& \ref{fig:shap}). 
In both cases, the resilience of most learning algorithms is sub-optimal and for several methods it drops with the correlation.

For the non-discriminatory case (i), \citet{Lipton2018Does} demonstrates that the algorithms fighting the disparities in treatment and impact~\cite{Pedreshi2008Discrimination, Feldman2014Certifying, Zafar2015Fairness} induce ``reverse'' discrimination. Our measurements of resilience and input influence captures this result and extend it to methods based on equalized odds and disparate mistreatment (the orange and brown lines in the left Figure~\ref{fig:corr} and orange line in Figures ~\ref{fig:shap}a), including methods equalizing overall misclassification rate, false negative rate, and related measures (Appendix D).
The only methods that do not bias the models in this scenario are: traditional supervised learning and the two methods that fall back to it if there is no direct discrimination in the data, i.e., the game-theoretic method based on envy-freeness (yellow line overlaps with the red line in the left Figure~\ref{fig:corr}) and the OIM.

For the discriminatory case (ii), we observe that with the growing correlation the resilience of the OIM stays high, whereas of three other algorithms decreases, suggesting that they induce discrimination via association~\cite{Wachter2019Affinity}, i.e., they replace the protected attribute with its proxy thus replicating ``redlining'', which causes a drop in resilience (e.g., the blue dotted line in the right Figure~\ref{fig:corr} \& in Figure~\ref{fig:shap}b). Therefore, it is not sufficient to simply drop the protected attribute in traditional learning.
Some methods perform poorly irrespective of the correlations, e.g., ``Hardt'', because it allows direct discrimination (orange lines in Figure~\ref{fig:corr} \& \ref{fig:shap}).
Overall, the two cases show that many learning algorithms induce discrimination or directly discriminate, i.e., they yield biased models by changing the impact of $\X$ on $\hat{\Y}$ or are directly impacted by~$\Z$.

%


\section{Evaluation on real-world datasets}
\label{sec:realdata}
In the synthetic settings, we experimented in an idealized environment where we had full information on the discriminatory concept shift and, therefore, knew the non-discriminatory ground truth. However, with real-world scenarios it is often the case that we only have access to a potentially discriminatory dataset without any information about the concept shift or we have a concept shift under a complex non-linear function.  Therefore, we analyze the OIM in two types of real-world settings. Firstly, on tabular datasets commonly found in algorithmic fairness research where we have multiple protected attributes and no information on the concept shift. Then, on the CelebA image dataset \cite{liu2015faceattributes} where we have non-discriminatory labels and introduce a discriminatory concept shift, while working with a highly non-linear deep neural net. 
\begin{figure}[tb]
\centering
\includegraphics[width=0.49\textwidth]{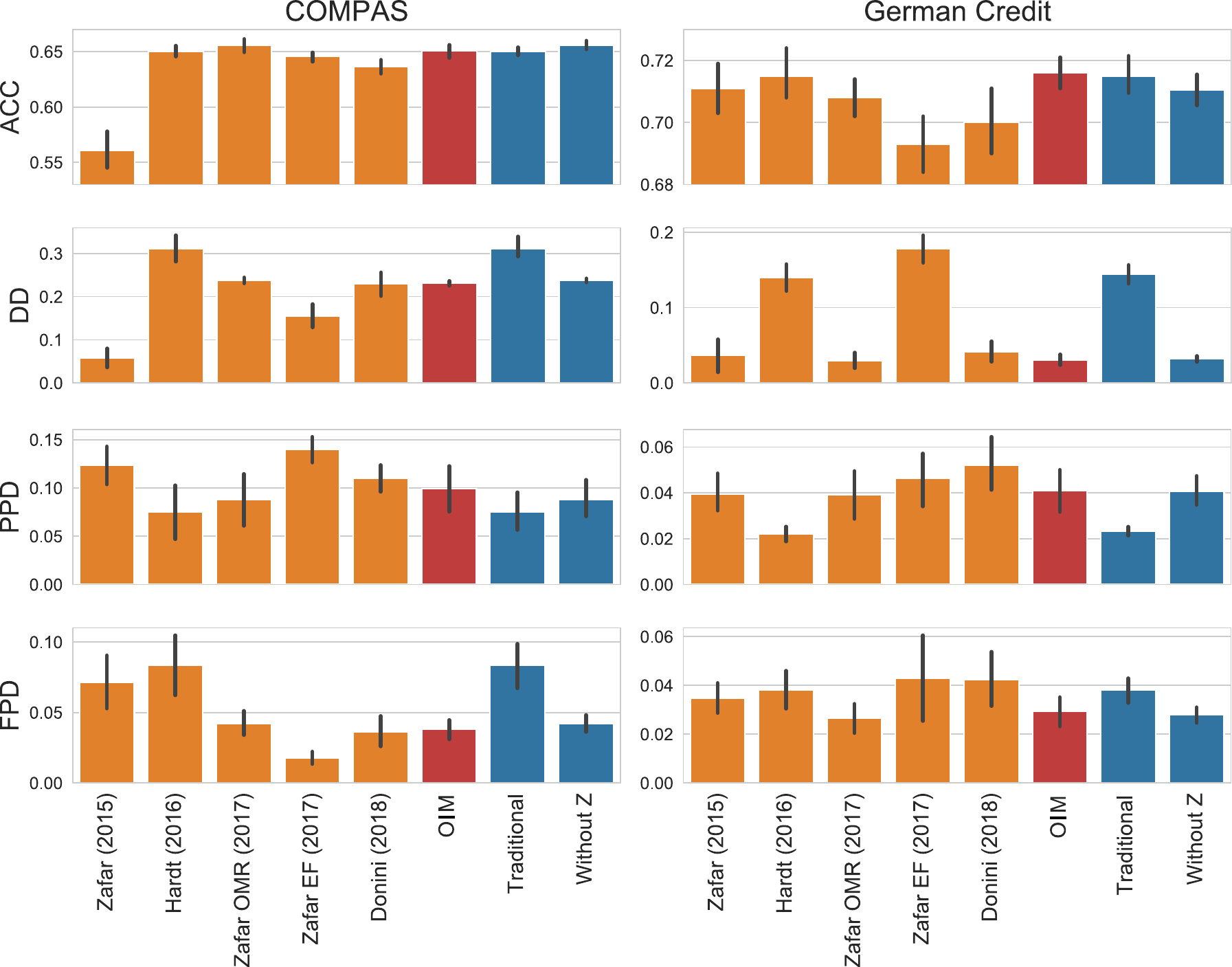}
\caption{Performance of learning algorithms inhibiting discrimination over COMPAS and German Credit datasets. Higher accuracy (ACC) and lower demographic disparity (DD), positive predictive disparity (PPD), and false positive disparity (FPD) are better.}\label{fig:grid}
\end{figure}

\begin{figure*}[tb]
\centering
  \begin{subfigure}[b]{0.24\linewidth}
    \includegraphics[width=\linewidth]{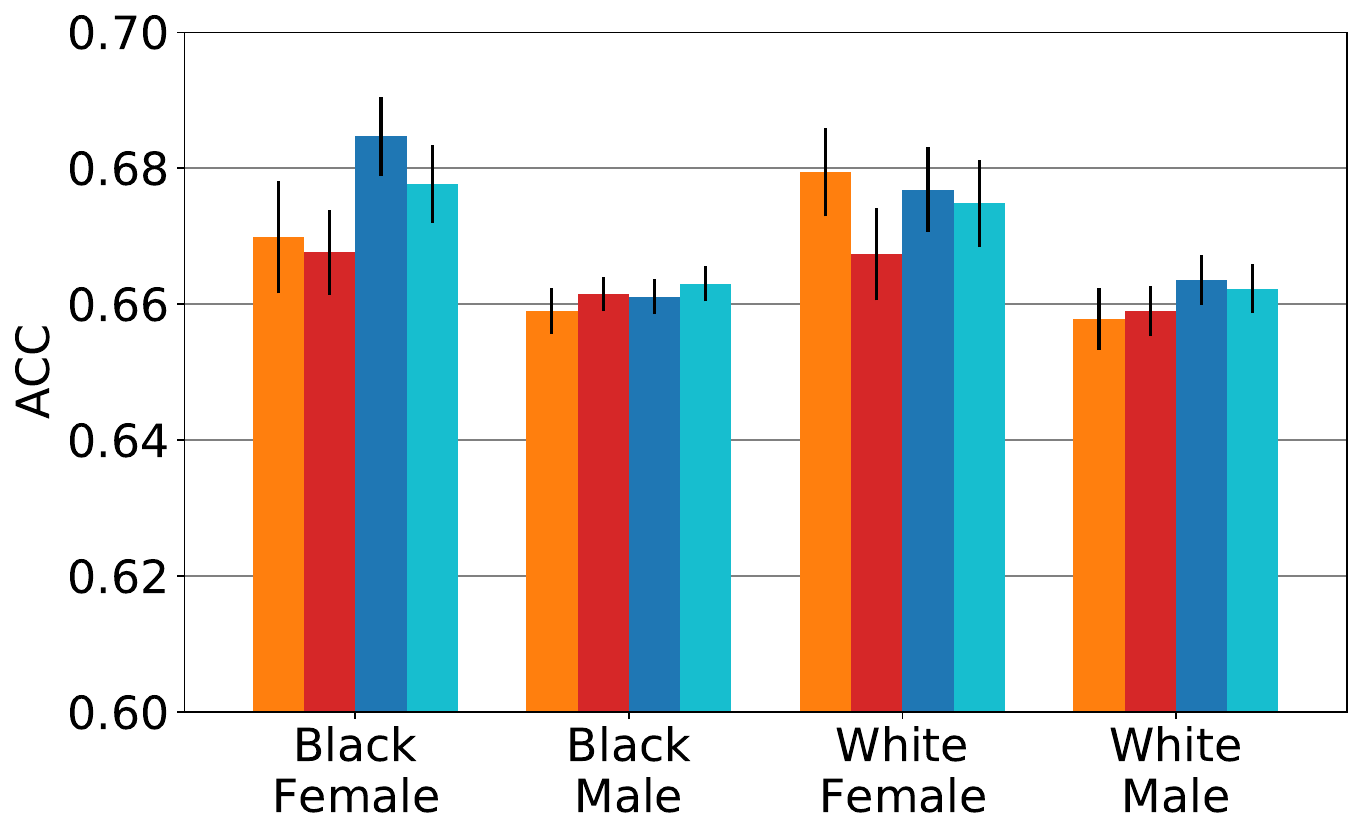}
  \end{subfigure}
  \begin{subfigure}[b]{0.24\linewidth}
    \includegraphics[width=\linewidth]{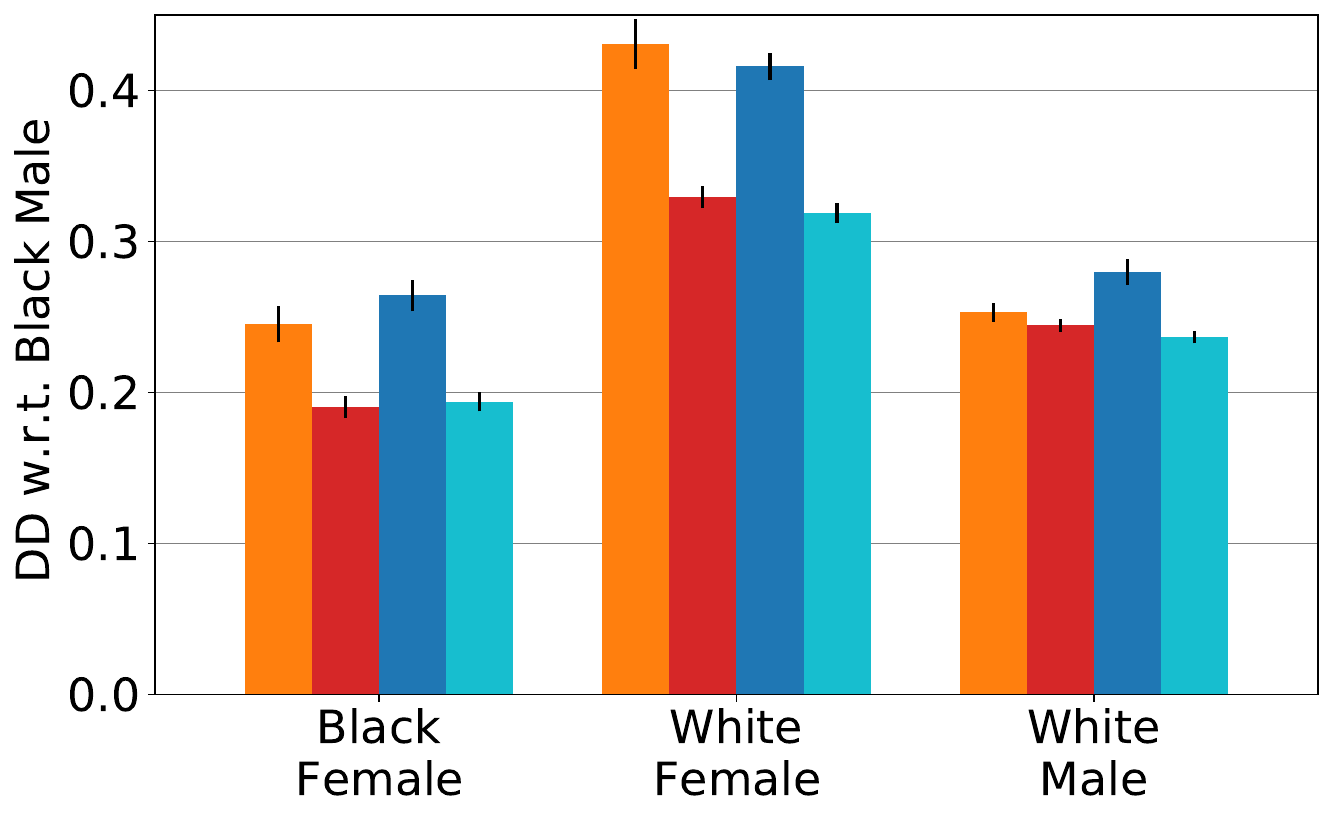}
  \end{subfigure}
  \begin{subfigure}[b]{0.24\linewidth}
    \includegraphics[width=\linewidth]{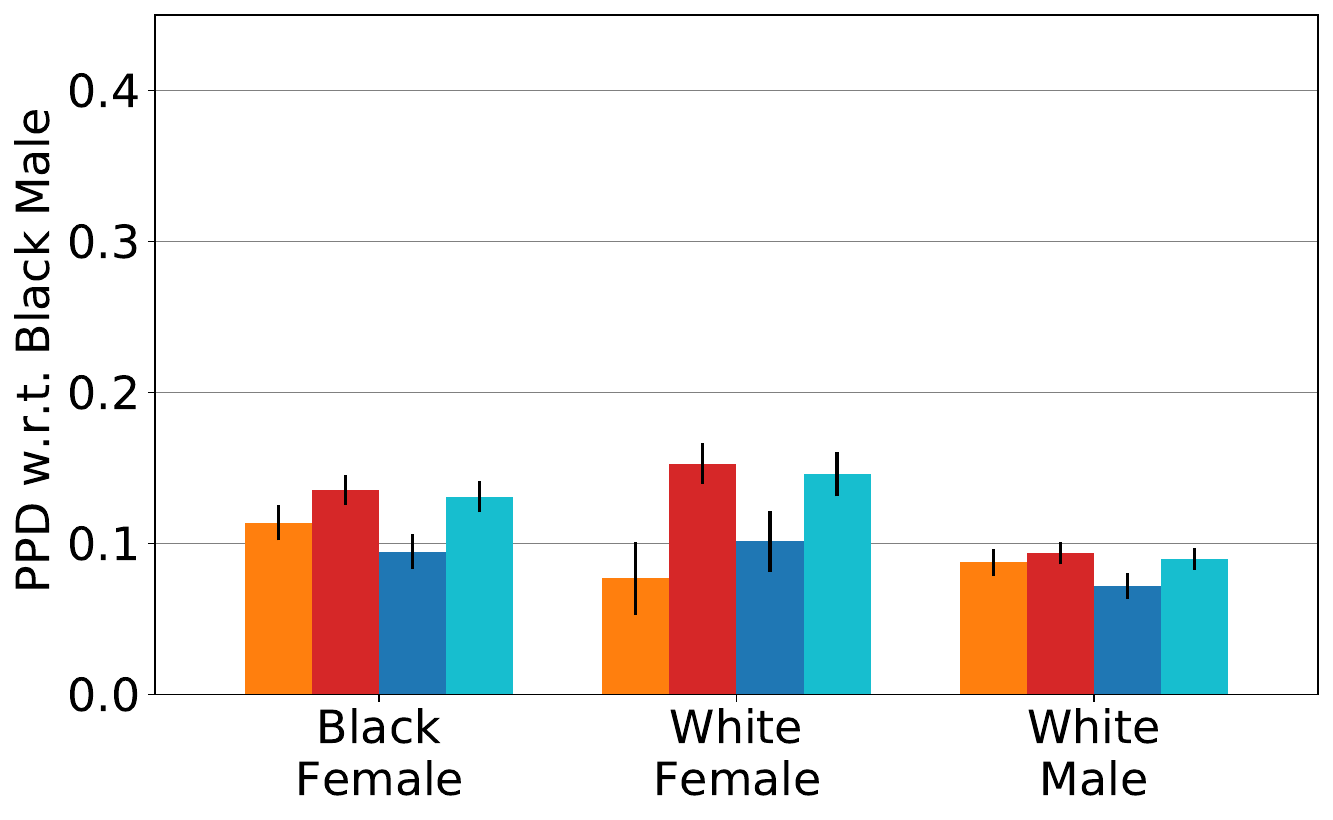}
  \end{subfigure}
  \begin{subfigure}[b]{0.24\linewidth}
    \includegraphics[width=\linewidth]{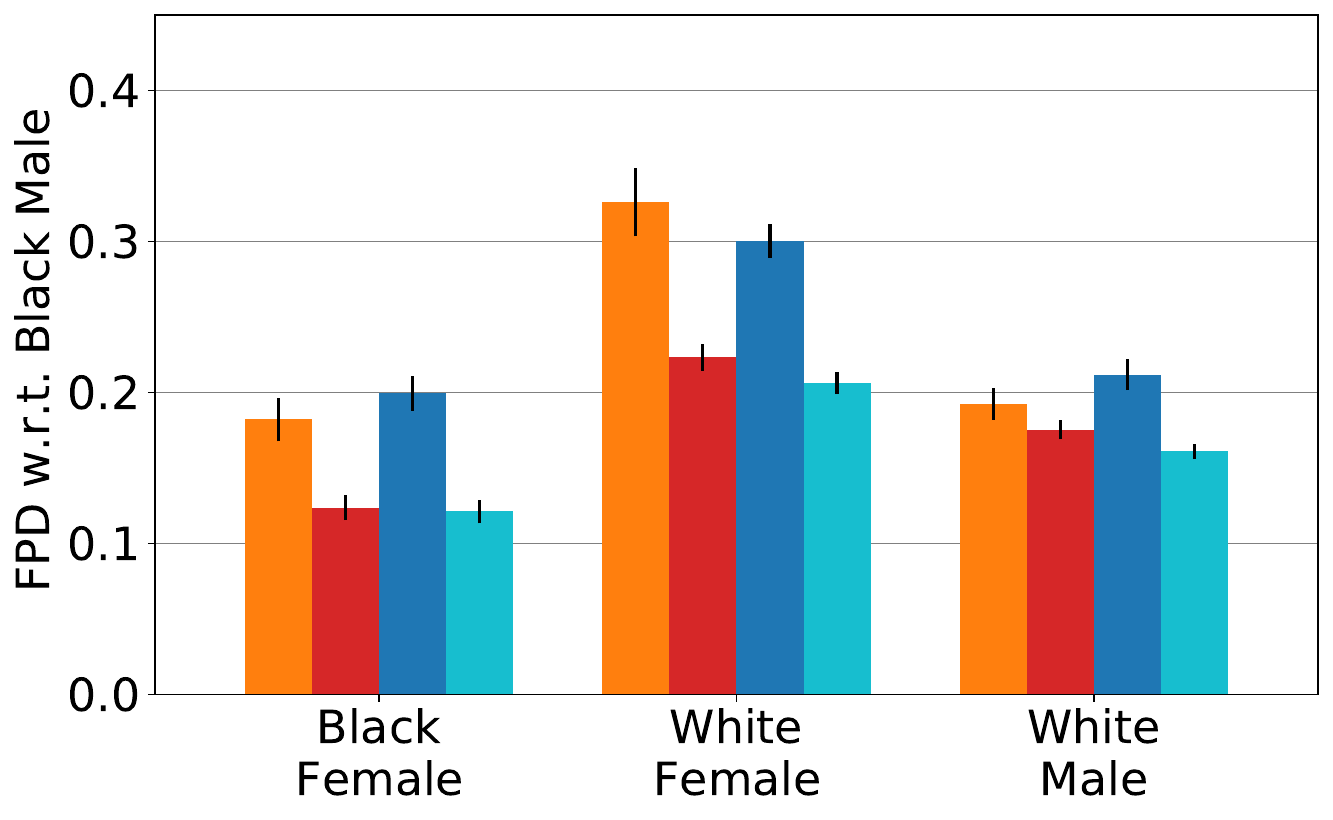}
  \end{subfigure}
  \begin{subfigure}[b]{0.5\linewidth}
    \includegraphics[width=\linewidth]{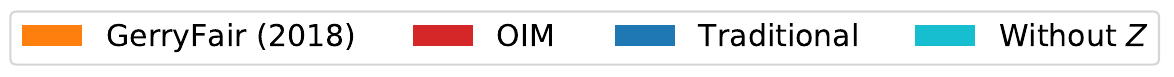}
  \end{subfigure}
\caption{Performance of learning algorithms inhibiting discrimination over all combinations of race \& sex on COMPAS. Disparity measures are on each given group w.r.t. Black Males. Higher accuracy (ACC) and lower disparities (DD, PPD, FPD) are better.}
\label{fig:multi}
\end{figure*}

\subsection{Concept shift information unknown}
\label{sec:tabular results}
\textbf{Datasets.}
We focus on two datasets that are prevalent in the literature on fairness: the COMPAS dataset of recidivism risk \cite{larson2016how} and the German Credit dataset of creditworthiness \cite{Dua:2019}, and their respective binary classification tasks. 

The ProPublica COMPAS dataset \cite{larson2016how} contains the records of 7214 offenders in Broward County, Florida in 2013 and 2014. As target, $y$, we use the binary label describing whether an individual recommitted a crime ($y= 1$). For comparison with the original study~\cite{larson2016how}, we follow their labeling of recidivism as the positive outcome.  In our single-protected attribute scenario we use the race (African American, Caucasian) as the protected feature, $Z$. We use race and sex (male, female) in the multiple protected attribute scenario. This dataset also includes information about the severity of charge, the number of prior crimes, and the age of individuals. 

The German Credit Dataset \cite{Dua:2019} provides information about 1000 individuals and the corresponding binary labels describing them as creditworthy ($y= 1$) or not ($y= 0$). Each variable $\x$ includes 20 attributes with both continuous and categorical data. We use the binary gender of individuals as the protected feature. This dataset also includes information aboutthe age, job type, housing type, and total amount in bank accounts of applicants and the total amount in credit, the duration, and the purpose of loan applications. 

\textbf{Measures.}
Since the non-discriminatory ground truth is unknown for these datasets, we use standard accuracy and demographic disparity to compare the learning algorithms. Demographic disparity measures disparate impact: $\text{DD}=|P(\hat{y}=1|z=0) - P(\hat{y}=1|z=1)|$~\cite{Zafar2015Fairness, Salimi2019Capuchin}. 
While other measures have been proposed and used in the  real-world context of applications~\cite{larson2016how}, such as disparity in false positive rate ($\text{FPD}=|P(\hat{y}=1|y=0, z=0) - P(\hat{y}=1|y=0, z=1)|$) or positive predictive value ($\text{PPD}=|P(y=1|\hat{y}=1, z=0) - P(y=1|\hat{y}=1, z=1)|$), both of which we report, these and other measures derived from the confusion matrix are determined by accuracy and demographic disparity~\cite{Narayanan2018Tutorial, Chouldechova2017Fair, Kleinberg2017Inherent, Friedler2016impossibility}. For the multiple protected attribute scenario, we report disparity for each combination of sex and race w.r.t. the largest and, across each measure, the most disadvantaged group in COMPAS, Black males.

\textbf{Results.}
We report the mean of the accuracy and disparities for the single-protected attribute scenarios and the multi-protected attribute COMPAS scenario in Figures \ref{fig:grid} \& \ref{fig:multi} respectively.

For the German Credit data, the OIM achieves the lowest demographic disparity and the highest accuracy (right panels of Figure~\ref{fig:grid}). For the COMPAS data on one protected attribute it also achieves the top accuracy, while yielding medium demographic disparity. The method that achieves much lower demographic disparity than the OIM directly constrains disparate impact at the expense of drastically lower accuracy and higher other disparities (''Zafar'' in the top left panel of Figure \ref{fig:grid}). The OIM also performs well in terms of false positive disparity and has medium performance for positive predictive disparity (four bottom panels in Figure \ref{fig:grid}). 

In the multiple protected attribute scenario, the OIM performed better than the traditional and the fair-learning method, ``GerryFair'' \cite{gerryfair}, in demographic and false positive disparities, while maintaining high accuracy (Figure \ref{fig:grid} \& \ref{fig:multi}). Therefore, the OIM addresses the substantial disparities in false positive rates by race reported in ProPublica's analysis of COMPAS over all intersections of race and sex \cite{larson2016how}. Even though the OIM resulted in marginally worse positive predictive disparity than the traditional method, as revealed in ProPublica's analysis and our results, this disparity is minimal to begin with. 
Note that tuning the ``GerryFair'' method's  parameters either increased accuracy with more disparity or vise-versa. 

In both datasets and protected attribute scenarios, the OIM performs similarly to the traditional method that drops the protected attributes, ``Without $Z$'', and select state-of-the-art methods; however, these methods does not offer any protections, nor guarantees, against induced discrimination, as described in \S\ref{sec:oip}, and for the other datasets we studied they induce discrimination and/or directly discriminate (see \S\ref{sec:synthetic} and \S\ref{sec:imageresults}). 

\subsection{Concept shift information known}
\label{sec:imageresults}

\begin{figure*}[t]
\centering
\begin{subfigure}[h]{0.20\linewidth}
    \includegraphics[width=.48\linewidth]{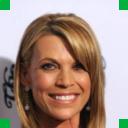}
    \includegraphics[width=.48\linewidth]{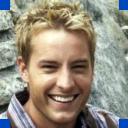}
    \caption{}
\end{subfigure}
\begin{subfigure}[h]{0.25\linewidth}
    \includegraphics[width=1\linewidth]{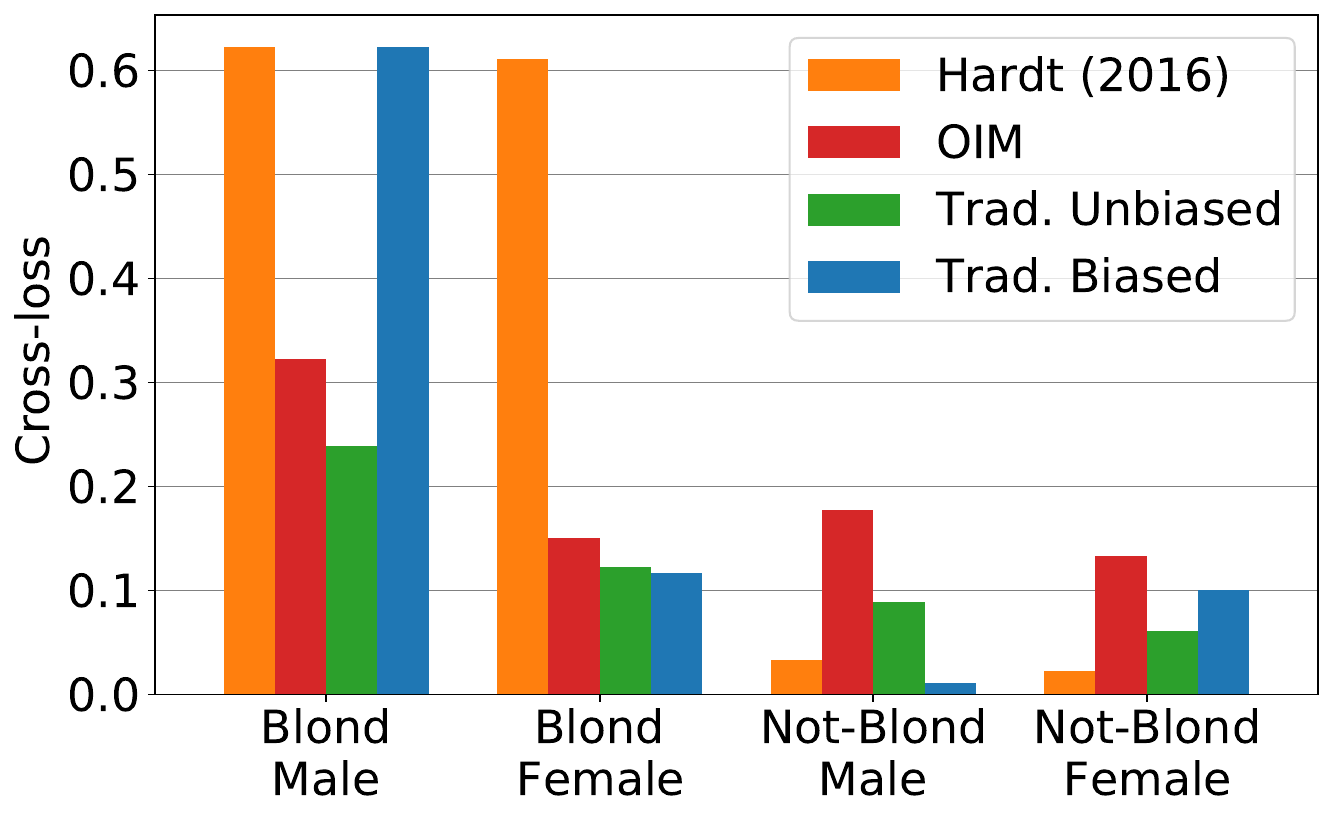}
    \caption{}
\end{subfigure}
\begin{subfigure}[h]{0.25\linewidth}
    \includegraphics[width=1\linewidth]{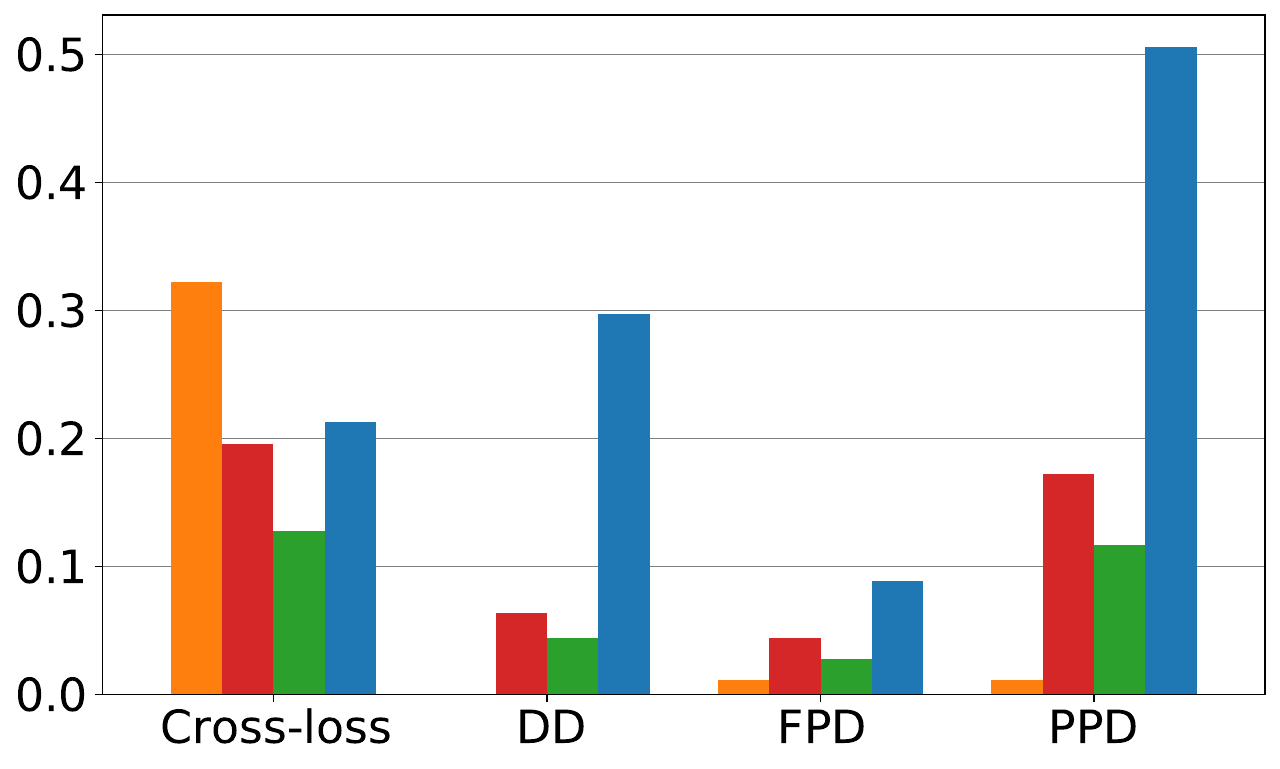}
    \caption{}
\end{subfigure}
\caption{
The expected cross-loss by hair-gender group (left plot) and the overall performance (right plot) of learning algorithms trained on the biased data following a discriminatory concept shift, except for the traditional trained on unbiased data (green bar). Marker style are shown in the photos on the left and have width of 10 pixels. Lower values are better. ``Traditional'' is ResNet-18. 
}
\label{fig:celeb}
\end{figure*}

\begin{figure}[t]
\centering
    \includegraphics[width=0.49\linewidth]{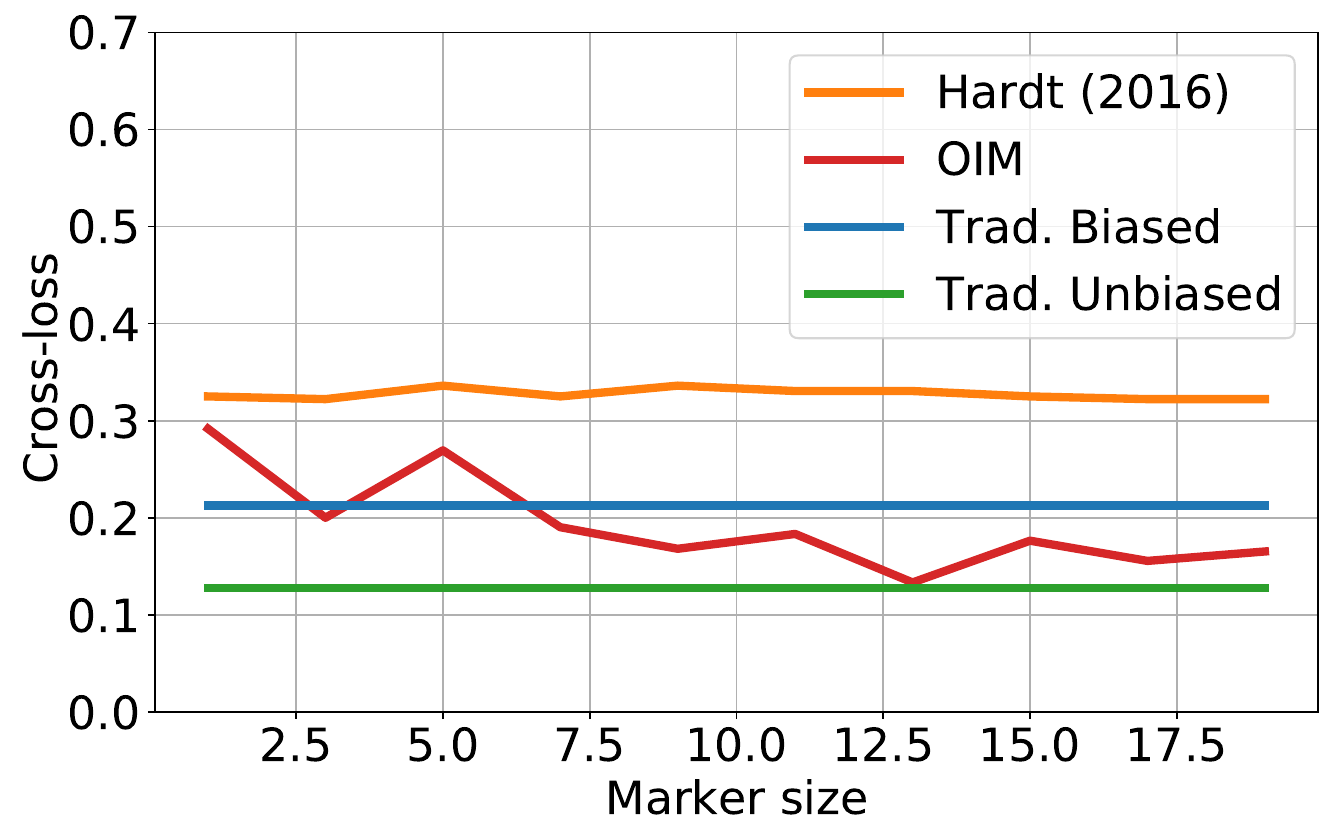}
    \includegraphics[width=0.49\linewidth]{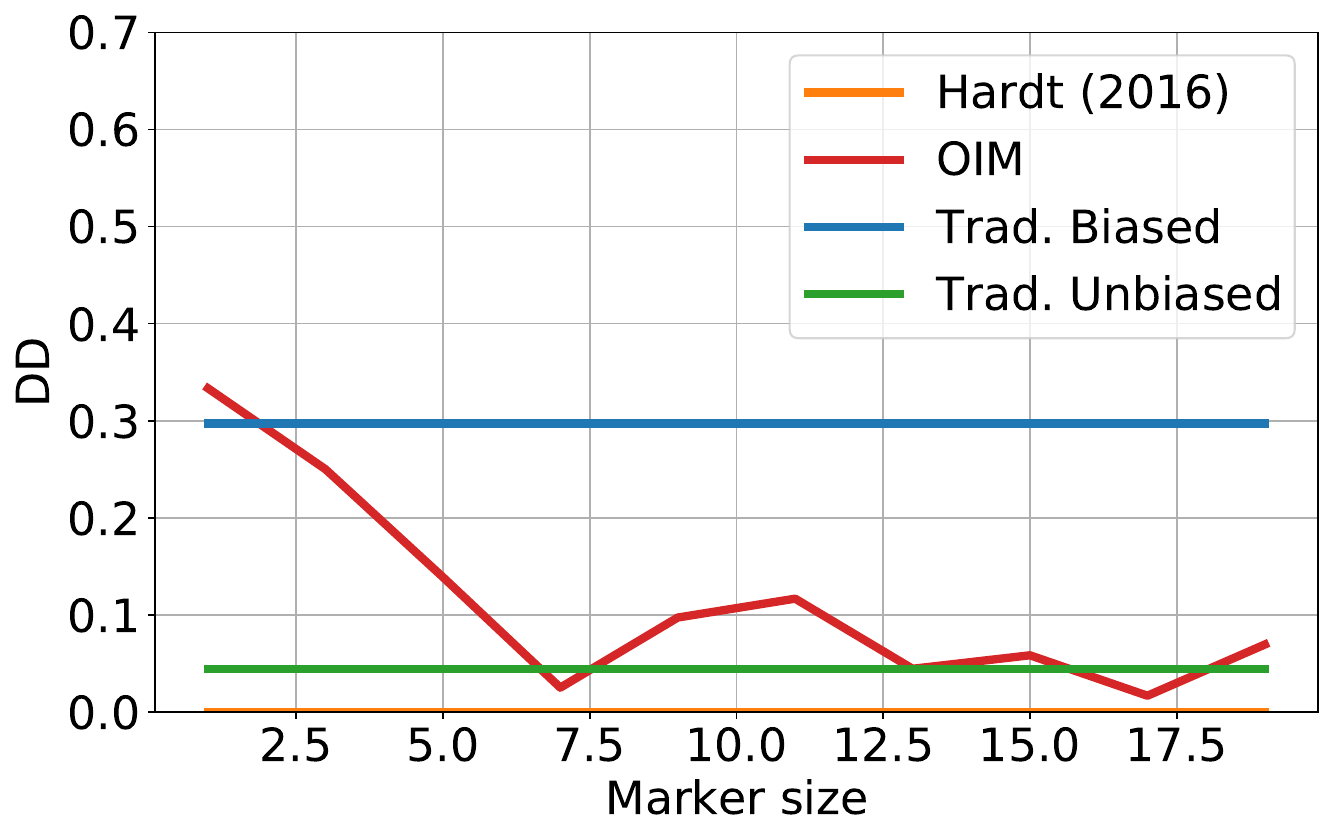}
\caption{
Overall expected cross-loss and demographic disparity of learning algorithms as marking pixel size increases. Marker style as in \ref{fig:celeb}a. Lower values are better. ``Traditional'' is ResNet-18. 
}
\vspace{-1px}
\label{fig:celebsize}
\end{figure}

\textbf{Dataset.}
We focus on the CelebA dataset \cite{liu2015faceattributes} commonly found in computer vision and deep learning literature. 
Here, the task is to classify the hair color of celebrities in photos, so the target labels are unlikely to be affected by any discrimination. That is, the non-discriminatory~$U$ is known and we can simulate discriminatory concept shift by swapping hair color labels to generate a discriminatory~$\Y$, which enable the measurements of cross-loss in real-world scenarios. 

CelebA is composed of celebrity images, each with 40 attribute annotations. Each image is transformed to 128*128 pixels, constituting the features $\X$.
We use the official train-val-test split from \citet{liu2015faceattributes} with blond ($y=1$) or not blond hair ($y=0$) as the target and binary gender as the protected attribute. To avoid sampling bias w.r.t. the hair-gender groups, we balance the dataset based on the smallest group (blond males). The balanced training and testing sets have 5,548 and 720 samples. To simulate a discriminatory concept shift, we randomly swap the labels of 50\% of blond males to not blond in the training data. We train the methods on this discriminatory data, except for the traditional method trained on the non-discriminatory data (green in Figure~\ref{fig:celeb}~\&~\ref{fig:celebsize}).

\textbf{Models and training.}
As our base model architecture we use a Pytorch implementation of ResNet-18~\cite{He2016DeepRL}. In addition to the OIM, only one of the evaluated learning methods' implementation, ~\citet{Hardt2016Equality}, can handle deep learning models, since both of them are post-processing methods. Therefore, all the methods train ResNet-18 on the images without annotations, then both fair learning methods use the gender annotations in their post-processing step.
The OIM also requires the addition of the protected attribute to the feature set when training ResNet-18. To avoid any changes to the architecture, we encode gender in the images via special markings (e.g., 10 pixel wide green and blue boxes shown in Figure \ref{fig:celeb}a). First, we train ResNet-18 on the photos with markings. Then, we estimate the optimal mixing distribution, $\pi^*$, on the training data. At the test time, we first compute the ResNet-18 predictions on the photos with either value of the gender mark, and then we average these predictions using the learned mixing distribution. Note that we do not use the ground-truth gender for making predictions in the test set, but rather the counterfactual values of the gender markings. Other methods train without these markings. 

\textbf{Results.}
We measure the expected cross-loss, demographic disparity (DD), false positive disparity (FPD), and positive predictive disparity (PPD). Despite training on the discriminatory data like the traditional biased method (blue in Figure \ref{fig:celeb}), the OIM reduces the expected cross-loss and the disparities close to that of the traditional unbiased method (red and green in Figure \ref{fig:celeb}). By contrast, when trained on discriminatory data, the traditional learning without $\Z$ (without markings) performs poorly both in terms of disparities and the cross-loss, especially for blond males whose label was swapped (blue in Figure \ref{fig:celeb}). Without the gender encoding, the model uses visual features of the images, such as hair and face shape, as proxies for gender. The method by \citet{Hardt2016Equality} results in the lowest DD and PPD (orange bars in Figure~\ref{fig:celeb}). However, it yields the highest expected cross-loss, in particular for the group with biased labels, i.e., blond males, and its female counterpart. In addition, this method tends to be further away (than the OIM) from the vanilla Resnet-18 training on the non-discriminatory data in terms of disparities.
The presented OIM results use 10 pixel wide green boxes on the corners of images of females with same sized blue markings on male pictures (Figure~\ref{fig:celeb}a). 
The results for similar markings as Figure~\ref{fig:celeb}a are nearly the same (Appendix I). 
The expected cross-loss and the disparities of the OIM initially decrease monotonically with the width of the markings (Figure~\ref{fig:celebsize}). At the width of about $10$ pixels this trend flattens, both in terms of expected cross-loss and disparities, suggesting that the markings are sufficiently large already for the model to use them. We note that, in real-world application domains where cross-loss cannot be measured, the size of markings can be established based on the disparity measures.

\section{Conclusion}

\textbf{Discussion.}
Our results shed a new light on the problem of discrimination prevention in supervised learning. 
First, we propose a new objective for discrimination prevention in supervised learning seeking methods that are resilient to discriminatory dataset shifts.
Dataset shifts clarify the dataset issues that can lead to discriminatory models. Different dataset shifts can be identified and tackled with different learning methods, so the remaining big question is whether these methods can be combined or are conflicting.

Second, we show that the optimal interventional mixtures do not produce ``reverse'' discrimination nor induce discrimination.
In the scenarios where training data is not discriminatory, the proposed learning method falls back to a traditional learning, and hence it is safer for general use than other approaches. 
While we do not provide resilience guarantees for discriminatory concept shifts with other perturbations than additive perturbations, to our knowledge this is the first study to provide such guarantees. 
Future research can study other dataset shifts to clarify the limits of this approach.

Third, we show that the proposed method is applicable to real-world settings with multiple protected groups and meets the explainability goal of removing their discriminatory impact, while remaining compatible with existing legal systems. The method provides a solution to the widely-discussed issue of protected groups' intersectionality and strikes a balance between protected groups, i.e., it does not correspond to affirmative actions advantageous to certain groups. The method overall is transparent and relatively easy to communicate to policymakers and courtroom officials.
%

\textbf{Limitations.}
We studied a variety of datasets and models, finding support for our methods, but a wider set of scenarios could be considered.
In future, discriminatory concept shifts could be measured via randomized human subject experiments or observational studies, and fair learning methods could be evaluated on resulting datasets and benchmarks. 
For instance, one could identify the groups of discriminating and fair members of hiring teams, as in our running Example, via population-level mixture models without identifying the individuals that belong to them~\cite{Grabowicz2018Experimental}. Then, mixture components could be used to simulate realistic discriminatory and fair decisions. Such evaluation techniques would facilitate the comparisons and bolster the credibility of fair learning methods.

All fairness objectives run the risk of being misused by practitioners to justify that their decision-making systems are fair. In any decision-making scenario, our method requires understanding whether the relationships in the causal model are fair and not. However, a practitioner may neglect the proper understanding of the causal processes and their fairness, e.g., they may overlook indirect discrimination \S\ref{sec:indirect-removal}. While our method will eliminate direct discrimination, it would not remove indirect discrimination, unless it is applied in an appropriate way. Thus, we emphasize the utmost importance of collaboration with domain experts to better understand the underlying causal process and their interpretation when applying our method and any other fair-learning methods in consequential decision-making systems.
\newpage
\bibliographystyle{ACM-Reference-Format}
\bibliography{jabref,manual_not_repeated}

\newpage
\section*{Appendix A: Proofs}

\begin{figure}[!b]
\begin{subfigure}[b]{0.45\linewidth}
    \centering
    \includegraphics[width=\linewidth]{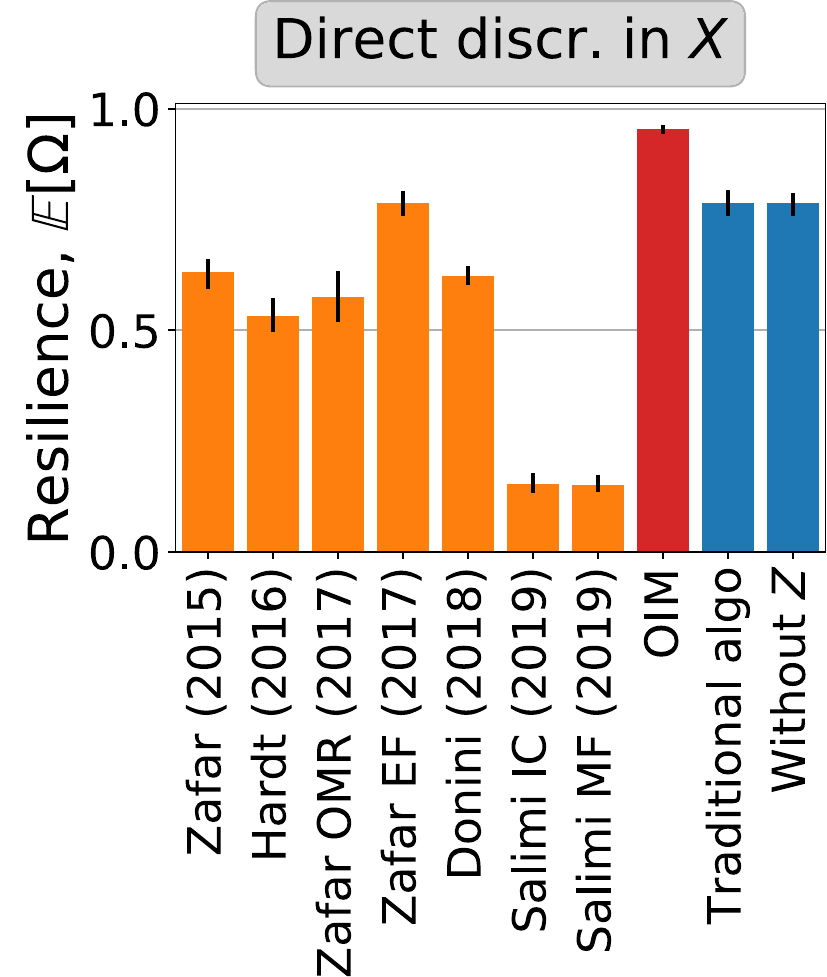}
    \caption{Logistic regression}
  \end{subfigure}
  \begin{subfigure}[b]{0.45\linewidth}
    \includegraphics[width=\linewidth]{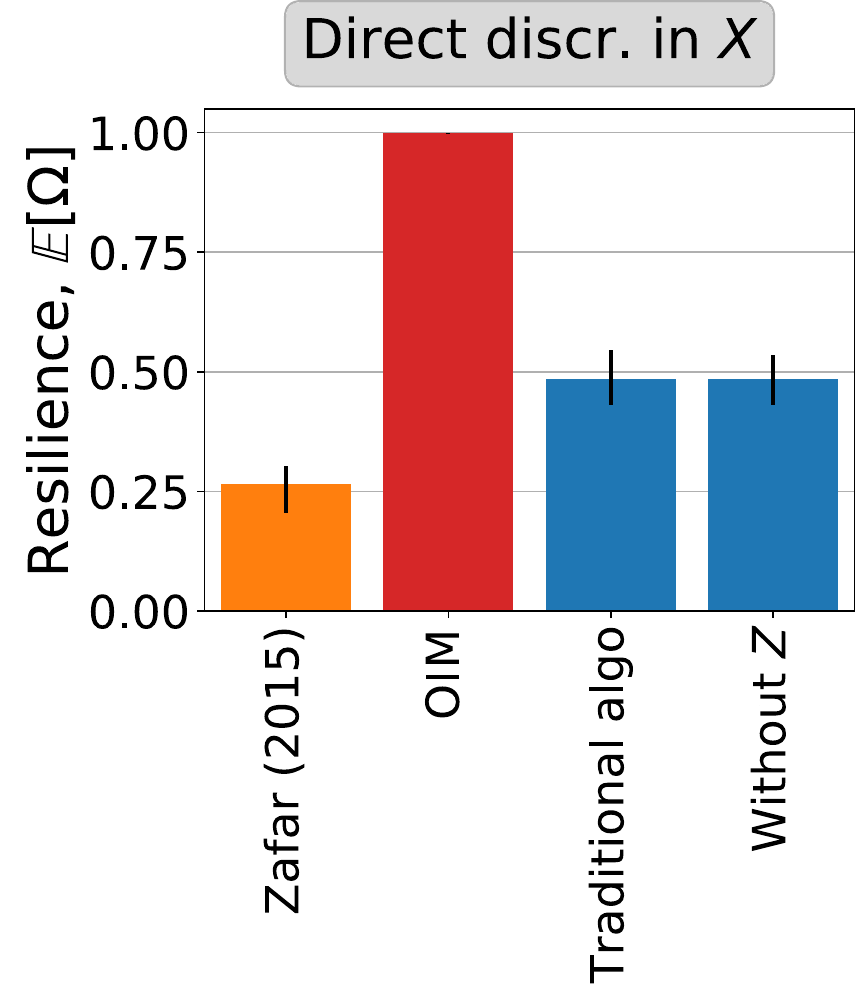} 
    \caption{Linear regression}
  \end{subfigure}
\caption{
Resilience of learning algorithms to discrimination in a relevant attribute ($\tilde{X}_1$) for logistic regression and linear regression.
}
\label{fig:disc_x}
\end{figure}

\begin{figure}[!b]
\centering
\begin{subfigure}[b]{0.95\linewidth}
    \includegraphics[width=\linewidth]{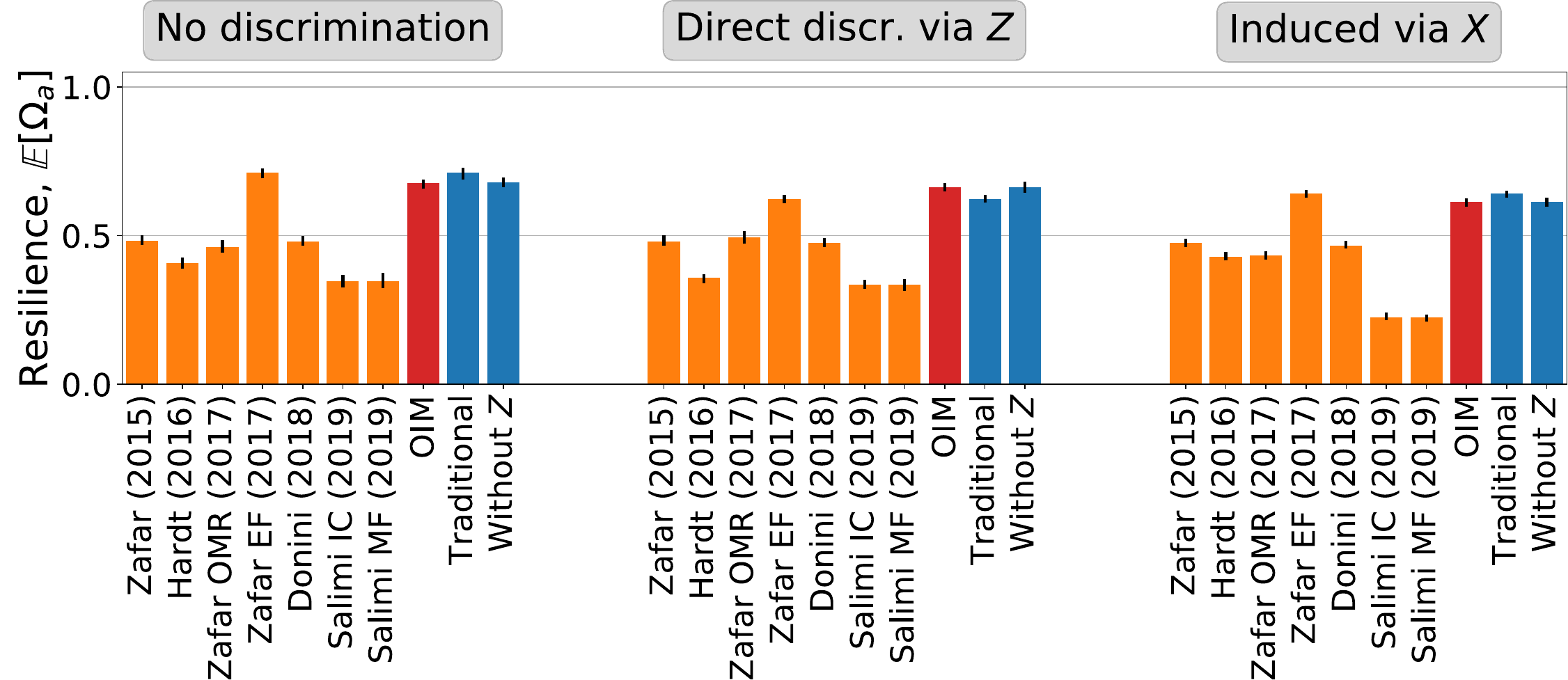}
    \caption{Logistic regression}
  \vspace{0.6cm}
  \end{subfigure}
  \begin{subfigure}[b]{0.95\linewidth}
    \includegraphics[width=\linewidth]{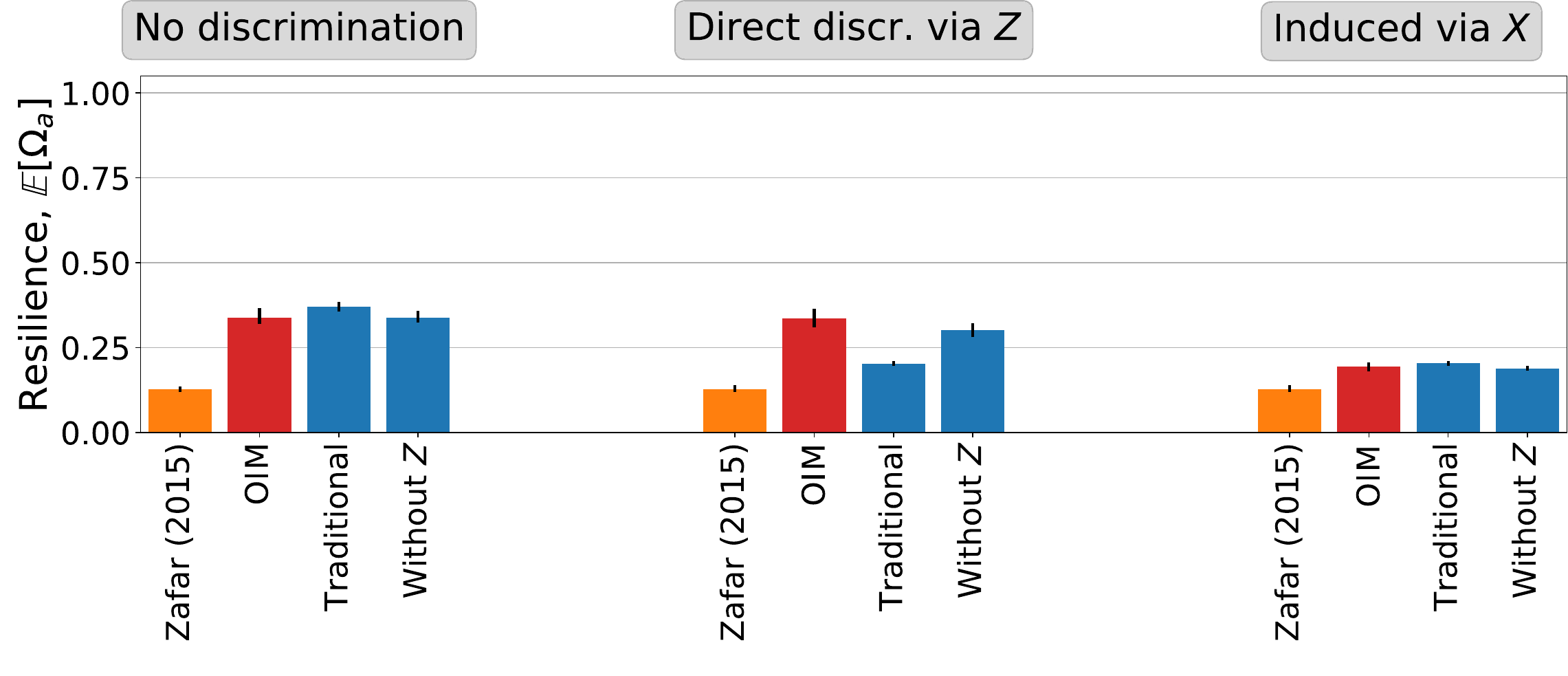}
    \caption{Linear regression}
  \end{subfigure}
\caption{
Resilience of learning algorithms with $X_2$ missing during training to non-discriminatory perturbations (the leftmost column) and discriminatory perturbation (the remaining two columns), for logistic regression and linear regression.
}
\label{fig:missing}
\end{figure}

\begin{figure*}[!ht]
\centering
\includegraphics[width=0.9\textwidth]{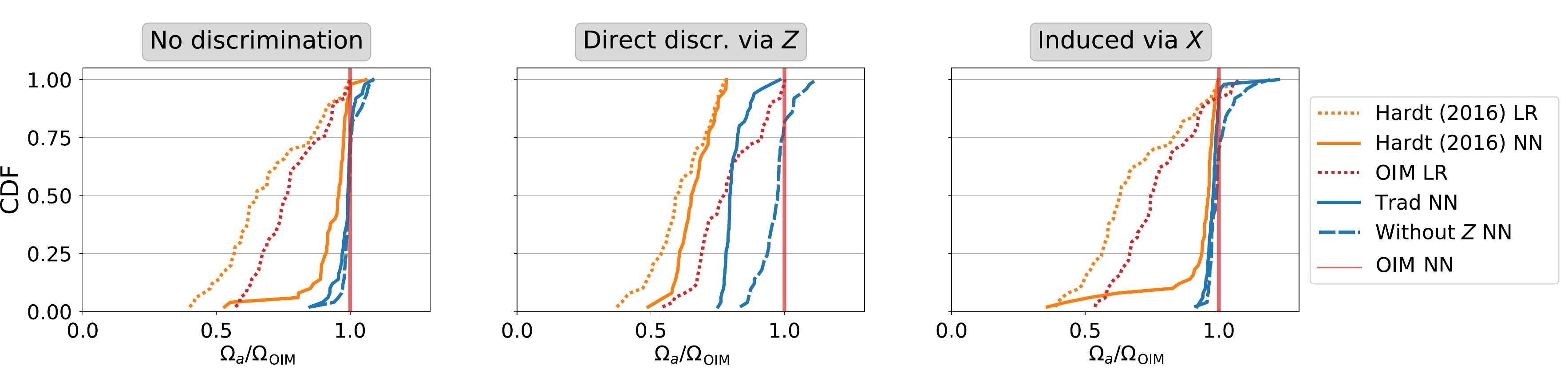}
\caption{Cumulative distribution function of per-dataset resilience of various learning algorithms~($\Omega_a$)
divided by the resilience of the optimal interventional mixture ($\Omega_{\text{OIM}}$)
for deep neural networks fitted to complex non-linear data generating models. 
The vertical red line is the CDF of the optimal interventional mixture applied to a neural network.
}
\label{fig:nn}
\end{figure*}

\begin{figure}[th]
\centering
\begin{subfigure}[h]{0.10\linewidth}
    \includegraphics[width=1\linewidth]{figs/0_g.jpg}\\
    \includegraphics[width=1\linewidth]{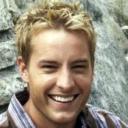}
    \caption{}
\end{subfigure}
\begin{subfigure}[h]{0.38\linewidth}
    \includegraphics[width=1\linewidth]{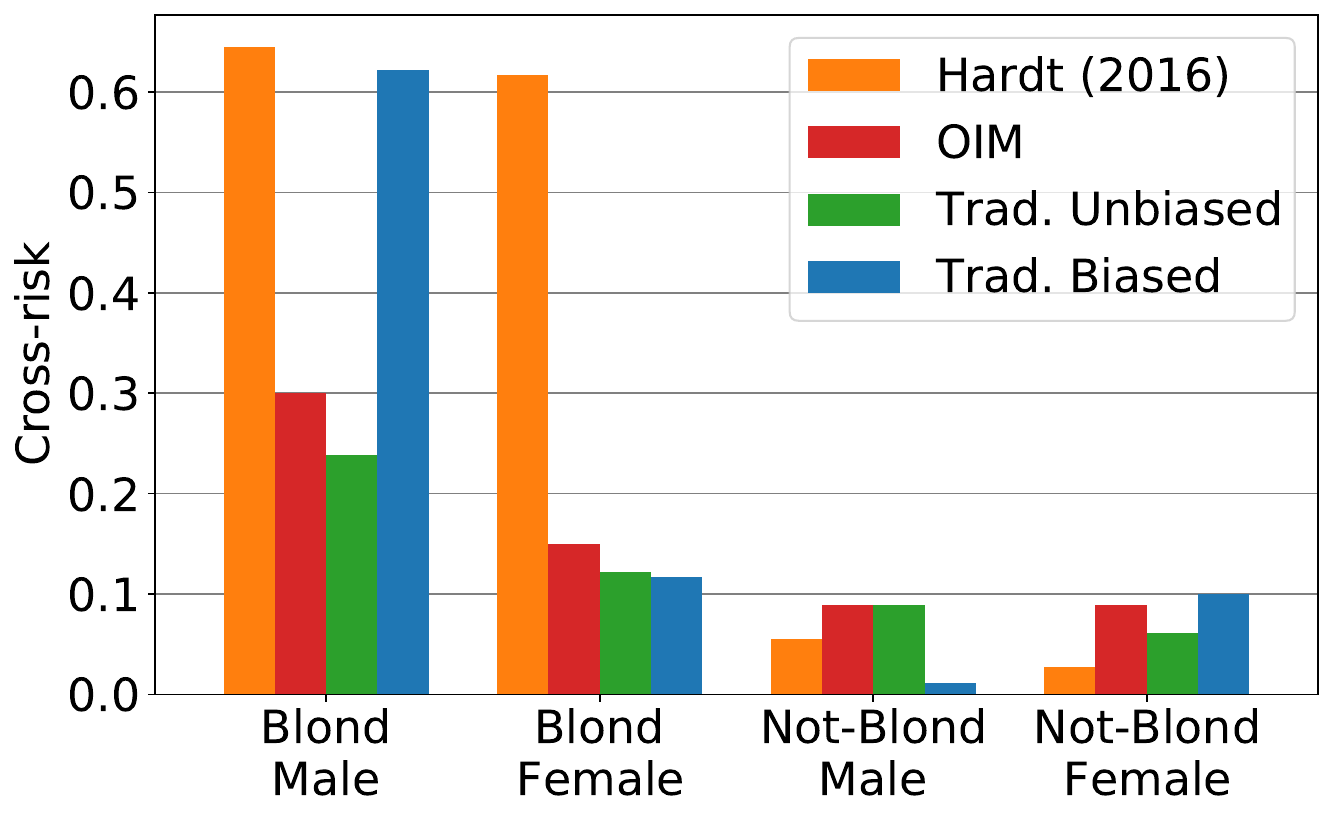}
    \caption{}
\end{subfigure}
\begin{subfigure}[h]{0.38\linewidth}
    \includegraphics[width=1\linewidth]{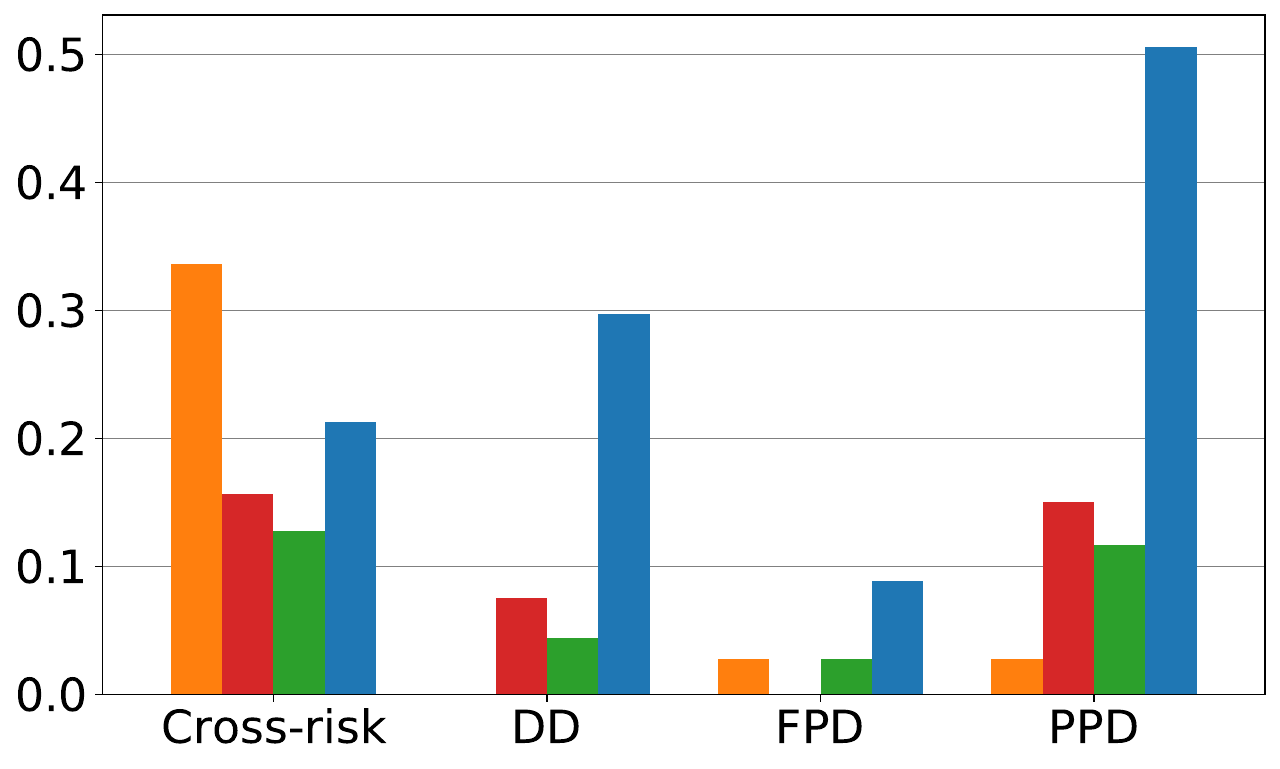}
    \caption{}
\end{subfigure}
\caption{
The cross-loss by hair-gender group (left plot) and the overall performance (right plot) of learning algorithms training on the perturbed data, except for one training on unbiased data (green bar). Marker style is as in (a) and size is 10 pixels. Lower values are better. ``Traditional'' is ResNet-18. 
}
\label{fig:celebaa}
\end{figure}

\begin{figure}[th]
\centering
    \includegraphics[width=0.49\linewidth]{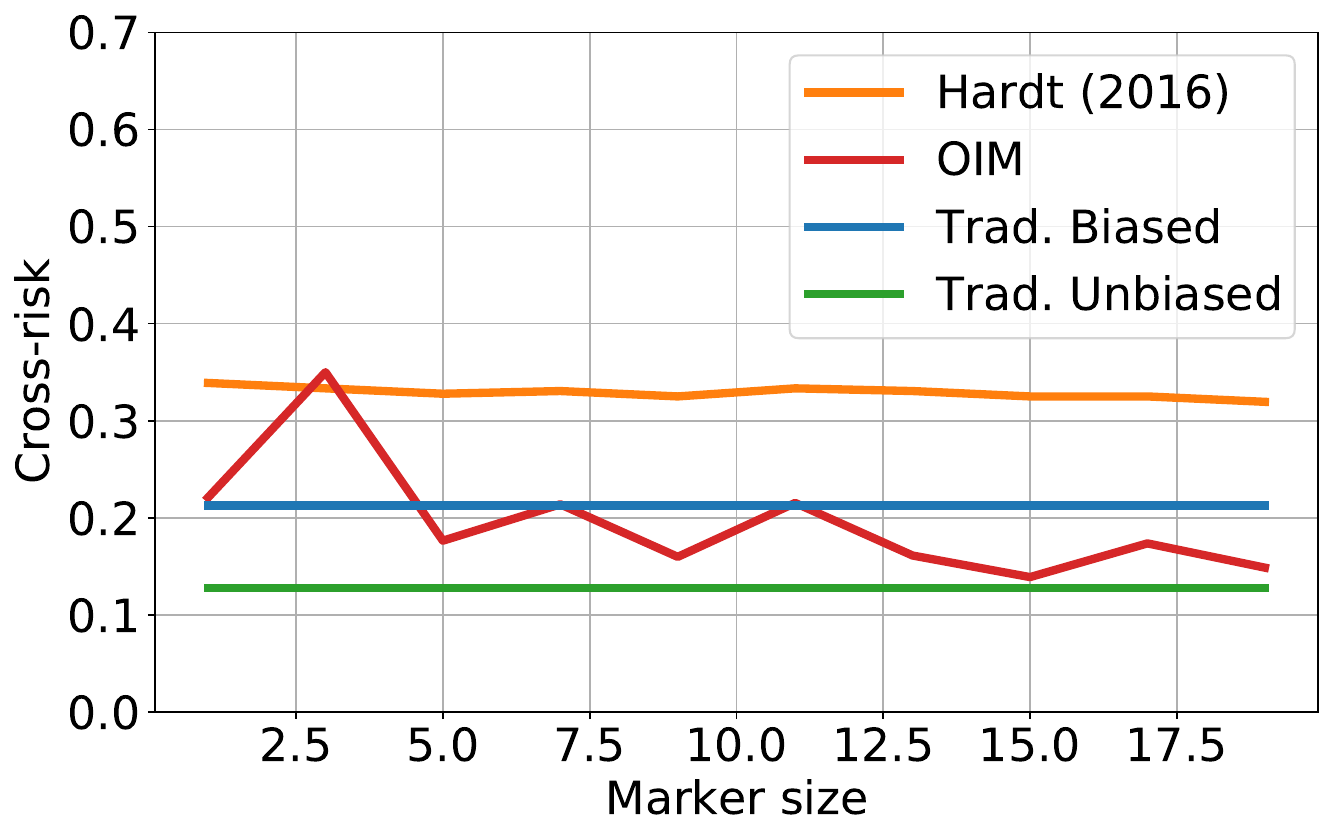}
    \includegraphics[width=0.49\linewidth]{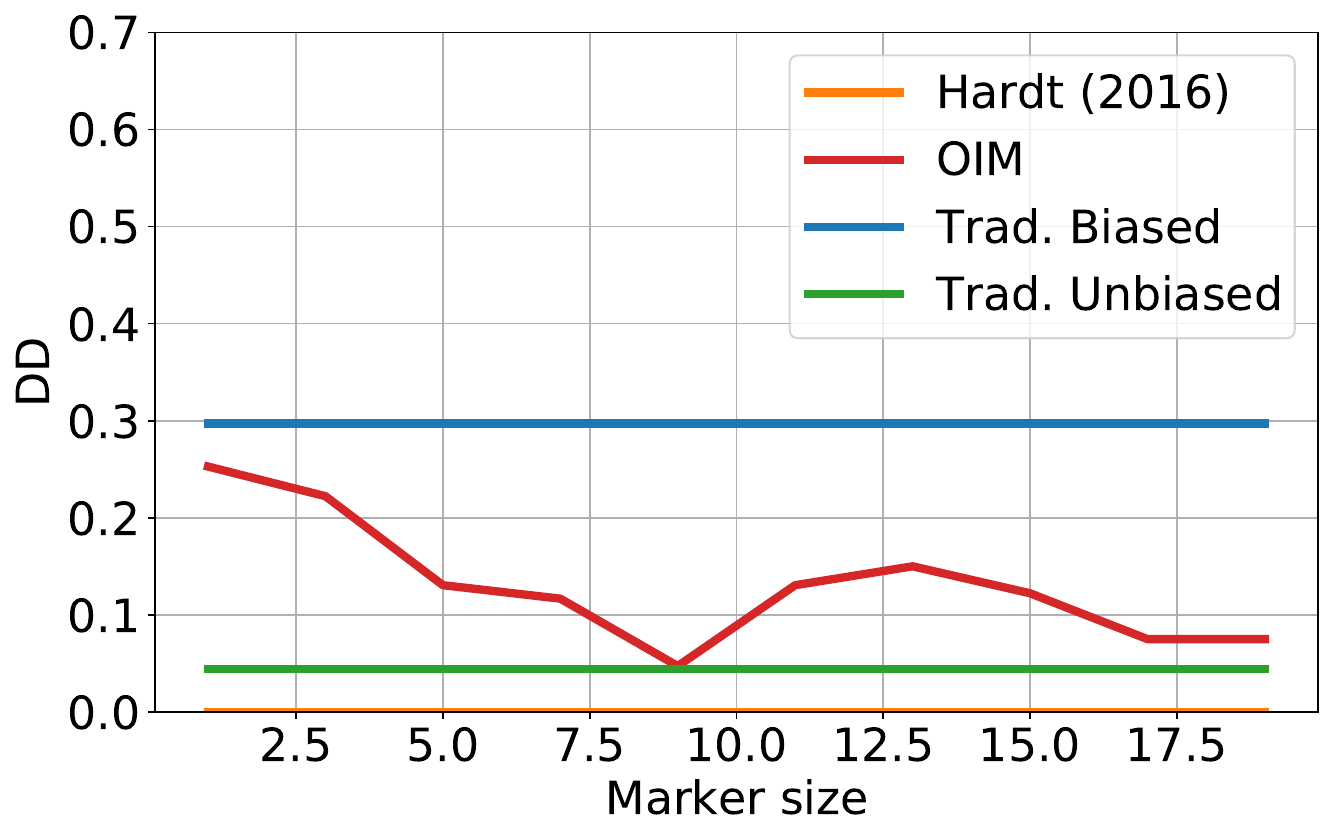}
\caption{
Overall cross-loss and demographic disparity of learning algorithms as marking pixel size increases. Marker style as in \ref{fig:celebaa}a. Lower values are better. ``Traditional'' is ResNet-18. 
}
\label{fig:celebsizesa}
\end{figure}

\textit{Proof of Proposition 1.}
From the definitions of consistent estimator and well-specified models, $\hat{\y}(\x,z) = \lim_{n\to\infty} \hat{\y}_n(\x,z) = f(\x) + h(z)$, where $n$ is the size of the training dataset, $\hat{\y}_n$ is a model trained on a given dataset.
Note that $\nu$ is centered at zero, e.g., $\E[\nu]=0$ under $\ell^2$ or $\mathbb{M}[\nu]=0$ under $\ell^1$, where $\mathbb{M}$ stands for median; otherwise $f(\x)$ can be redefined to center $\nu$. 
From the definition of the OIM and consistent estimator, $\hat{\y}^*(\x) = \lim_{n\to\infty} \hat{\y}^*_n(\x) = \E[\hat{\y}(\x,Z')] = f(\x)+C_p$.
For $\ell^2$ loss, $C_2=\E[h(Z)+\nu]$, while for $\ell^1$ loss, $C_1 = \mathbb{M}[h(Z)+\nu]$.
For given datasets $D$ and $\tilde{D}$, the smaller the denominator in the definition of resilience, $\E_D\left[ \ell\left( \U, \hat{y}_a( \X | \tilde{D} )\right) \right]$, the larger the resilience of the learning method.
For the OIM, the denominator is $\E_D\left[ \ell\left( U,\hat{\y}^*(\X)\right) \right] = \E[\ell(f(\X)+\nu,f(\X) + C_p )] = \E[|\nu- C_p|^p]$. 
If $\E[h(Z)]=0$ under $\ell^2$ loss or $\mathbb{M}[h(Z)+\nu] = \mathbb{M}[\nu]=0$ under $\ell^1$ loss, then the OIM strictly maximizes the resilience, achieving $\E_D\left[ \ell\left( U,\hat{\y}^*(\X)\right) \right]=0$ and $\Omega=1$.
For an arbitrary model $\hat{y}(\x)$, $\E_D\left[ \ell\left( \U, \hat{y}_a( \x | \tilde{D} )\right) \right]=\E[\ell(f(\X)+\nu,\hat{y}(\x) )] = 
\E_{\X} \E_{\nu|\X} [|\nu+ f_1(\X)|^p]\geq \E_{\X} \E_{\nu|\X} [|\nu|^p]=
\E_D\left[ \ell\left( U,\hat{\y}^*(\X)\right) \right]|_{C_p=0}$, where $f_1(\x)=f(\x)-\hat{y}(\x)$. Thus, the expected loss is minimized for $f_1(\X)=0 \iff \hat{y}(\x)=f(\x)$.


\textit{Proof of Corollary 1.}
Universal approximation theorems~\cite{Cybenko1989Approximation, Pinkus1999Approximation}, which show that the loss of a universal approximator is bounded, $\sup_{\x,z} \ell( g(\x,z), \hat{y}_\text{nn}(\x,z) )<\epsilon$, for any positive $\epsilon$ and any function $g(\x,z)$. In particular, $\ell^p(f(\x)+h(z), \hat{y}_\text{nn}(\x,z)) < \epsilon$ and $ < f(\x)+h(z) - \epsilon^p < \hat{y}_\text{nn}(\x,z) < f(\x)+h(z) + \epsilon^p$. From the definition of the OIM, we get $ < f(\x)+C_p - \epsilon^p < \hat{y}_\text{nn}(\x,z) < f(\x)+C_p + \epsilon^p$ and $\ell^p(f(\x)+h(z), f(\x)+C_p) < \epsilon$.

\textit{Proof of Proposition 2.}
Let the definitions and assumptions hold from the \textit{Proof of Proposition 1}.  
From the definition of the MIM and consistent estimator, $\hat{\y}_\text{MIM}(\x) = \lim_{n\to\infty} \hat{\y}^*_n(\x) = \E_{\Z'}[\hat{\y}(\x,Z')] = f(\x)+\E[h(Z)+\nu]$ for any loss. 
For given datasets $D$ and $\tilde{D}$, the smaller the denominator in the definition of resilience, \\$\E_D\left[ \ell\left( \U, \hat{y}_a( \X | \tilde{D} )\right) \right]$, the larger the resilience of the learning method.
For the MIM, the denominator is $\E_D\left[ \ell\left( U,\hat{\y}_{\text{MIM}}(\X)\right) \right] = \E[\ell(f(\X)+\nu,f(\X) +\E[h(Z)+\nu)] = \E[|\nu-\E[h(Z)]|^p]$. 
If $\E[h(Z)]=0$, then only under $\ell^2$ loss can the MIM strictly maximizes the resilience, achieving $\E_D\left[ \ell\left( U,\hat{\y}_{\text{MIM}}(\X)\right) \right] =0$ and $\Omega=1$.

\section*{Appendix B: Evaluated Methods' Parameter Choice}
For the method by \citet{Donini2018Empirical} we report the performance with a linear-kernel SVM; the regularization parameter $C$ for was tuned via grid search with $C \in \{0.01,0.1,1\}$. For \citet{Zafar2017Fairness} we report results for when the model is set to equalize misclassification rates between two groups. For \citet{Zafar2015Fairness} we set the constraint $c=0$. 
The only fair-learning method we evaluate in the multiple protected attribute setting is the method introduced in the fairness gerrymandering paper \cite{gerryfair}. For this method, we chose $\gamma= 0.3$, which resulted in an accuracy within a few percentile of traditional learning.

\section*{Appendix C: Data generation for random generalized linear models}

We generate a synthetic set of $10\,000$ samples $\{(\x,z)\}$ from a standard multivariate normal distribution with a random correlation matrix~\cite{Ghosh2003Behavior}. For simplicity, in our experiments we use two relevant features, that is $\x$ has two dimensions. The variable $z$ is converted to a binary value with the sign function.
The coefficients $\bm{\alpha}$, $\bm{\tilde{\alpha}}$, and $\beta$ are drawn from $\text{Uniform}[-5,5]$, unless specified otherwise.
We generate the non-discriminatory ground truth decisions, either as samples from 0-1 coin tosses, $\u \sim \text{Bernoulli}[\E[\U|\x]]$, or normal distribution with unit variance, $\u \sim \text{Normal}[\E[\U|\x],1]$.
The resulting set of samples constitute the unperturbed evaluation dataset $D=\{(\x,z,u)\}$.
Finally, we sample the perturbed decisions, $y \sim \p(\y|\x,z)$, which contribute to the training dataset $\tilde{D}=\{(\x,z,y)\}$. 

\section*{Appendix D: Evaluation on a hiring scenario}


Here, we present the results from a synthetic scenario proposed by \cite{Lipton2018Does}, modified slightly as follows. Using this example, we show how state-of-the-art learning algorithms addressing discrimination induce it even when the training data is non-discriminatory. 

To this end, we sample 1000 observations from the data-generating process below:
\begin{gather*}
    z_i \sim \text{Bernoulli}[0.5]\\
    \text{hair\_length}_i | z_i = 0 \sim  35 * \text{Beta[2, 2]}\\
    \text{hair\_length}_i | z_i = 1 \sim  35 * \text{Beta[2, 7]}\\
    \text{work\_exp}_i | z_i = 0 \sim \text{Poisson}[25] - \text{Normal}[20, \sigma = 0.2]\\
    \text{work\_exp}_i | z_i = 1 \sim 
\begin{cases}
    \text{Normal}[10, \sigma = 2]& \text{w/ prob 0.2}\\
    \text{Normal}[15, \sigma = 2]& \text{w/ prob 0.8}
\end{cases} \\
    p_i= f(-25.5 + 2.5*\text{work\_exp}) \text{ where } f(x) = \frac{1}{1+e^{-x}} \\
    y_i | \text{work\_exp} \sim \text{Bernoulli}[p_i]
\end{gather*}
This synthetic data represents the historical hiring process where the protected attribute is a candidate's gender, $z$. The data has the following properties: i) the hiring decision has been made based on the work experience only, thus, it is non-discriminatory data;  ii) since women in this scenario on average have less work experience than men, men have been hired at higher rate than women historically; and iii) women tend to have longer hair than men. Therefore, a model that uses hair length in its decision-making can induce indirect discrimination. Additionally, we introduced modifications to this synthetic data with respect to the original scenario~\cite{Lipton2018Does}. The work experience of male candidates now follows a bi-modal distribution (i.e., a mixture of two normal distributions) with one peak at 10 and another at 15. We trained a method for discrimination prevention \cite{Zafar2017Fairness} under three different fairness constraints: equalized \textit{misclassification rate}, \textit{false positive rate} (FPR), \textit{false negative rate} (FNR). We also trained a model while simultaneously optimizing both FPR and FNR; however, the learned model returned trivial predictions where all candidates are rejected.
The relative utility of the various methods is low compared with the OIM.

\begin{table}[ht]
\begin{center}
\centering
\begin{tabular}{l | c} 
 Method & $\E[R_\text{perf}/R]$  \\
 \hline
 OIM& 1.000  \\ 
 \citet{Zafar2017Parity} & 0.997 \\
 \citet{Zafar2017Fairness} with FNR & 0.838 \\
 \citet{Zafar2017Fairness} with Missclass. & 0.777 \\
 \citet{Donini2018Empirical} & 0.634 \\
 \citet{Zafar2017Fairness} with FPR & 0.570  \\
 \citet{Hardt2016Equality} & 0.328 \\
 \citet{Zafar2015Fairness}& 0.179 \\
\end{tabular}
\par\bigskip
\par\bigskip
\caption{Relative utility of various fairness models \cite{Hardt2016Equality, Donini2018Empirical, Zafar2015Fairness, Zafar2017Fairness, Zafar2017Parity} trained with the synthetic data}
\label{tab:results}
\end{center}
\end{table}

%

\section*{Appendix E: Random generalized linear models}
\begin{figure}[tb]
\centering
  \begin{subfigure}[b]{0.95\linewidth}
    \includegraphics[width=\linewidth]{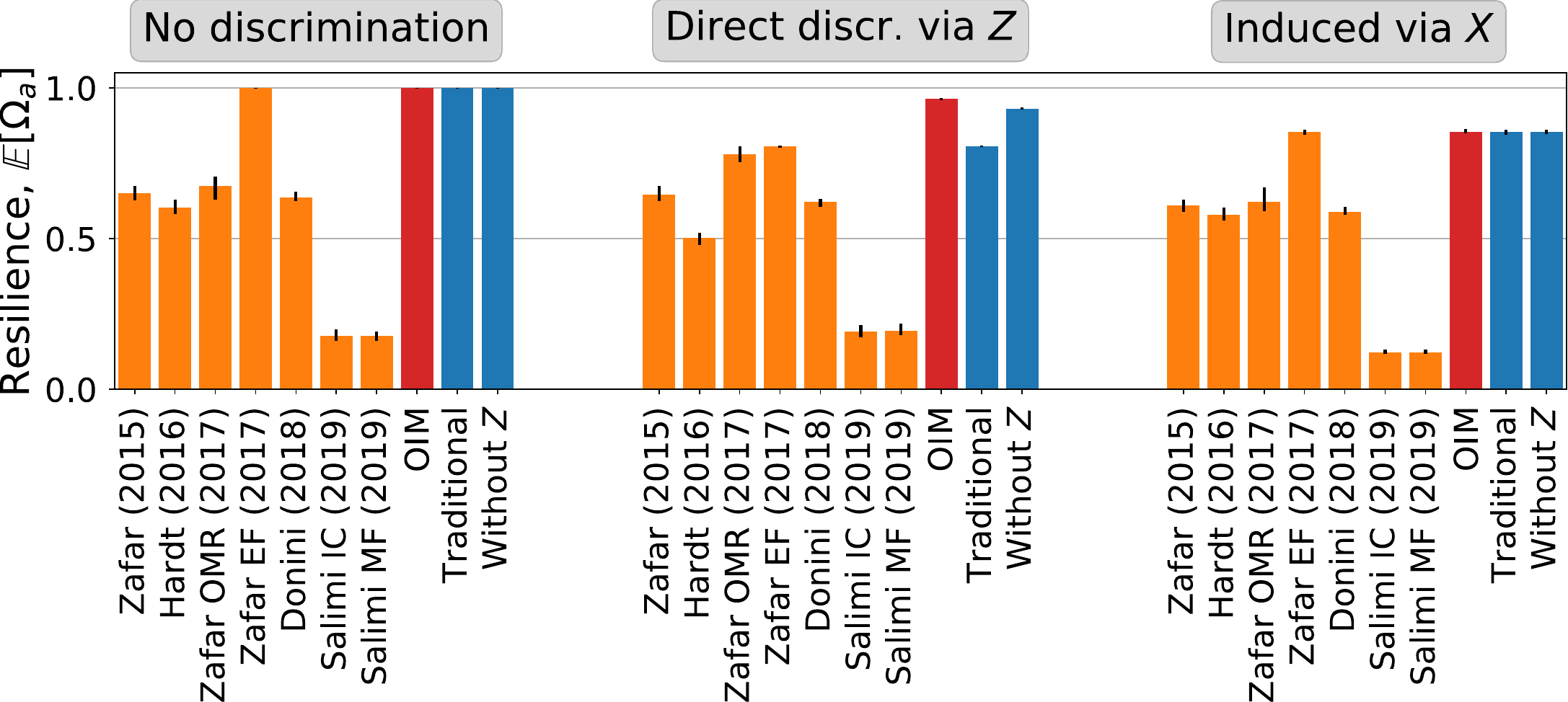}
    \caption{Logistic regression}
  \vspace{0.7cm}
  \end{subfigure}
  \begin{subfigure}[b]{0.95\linewidth}
    \includegraphics[width=\linewidth]{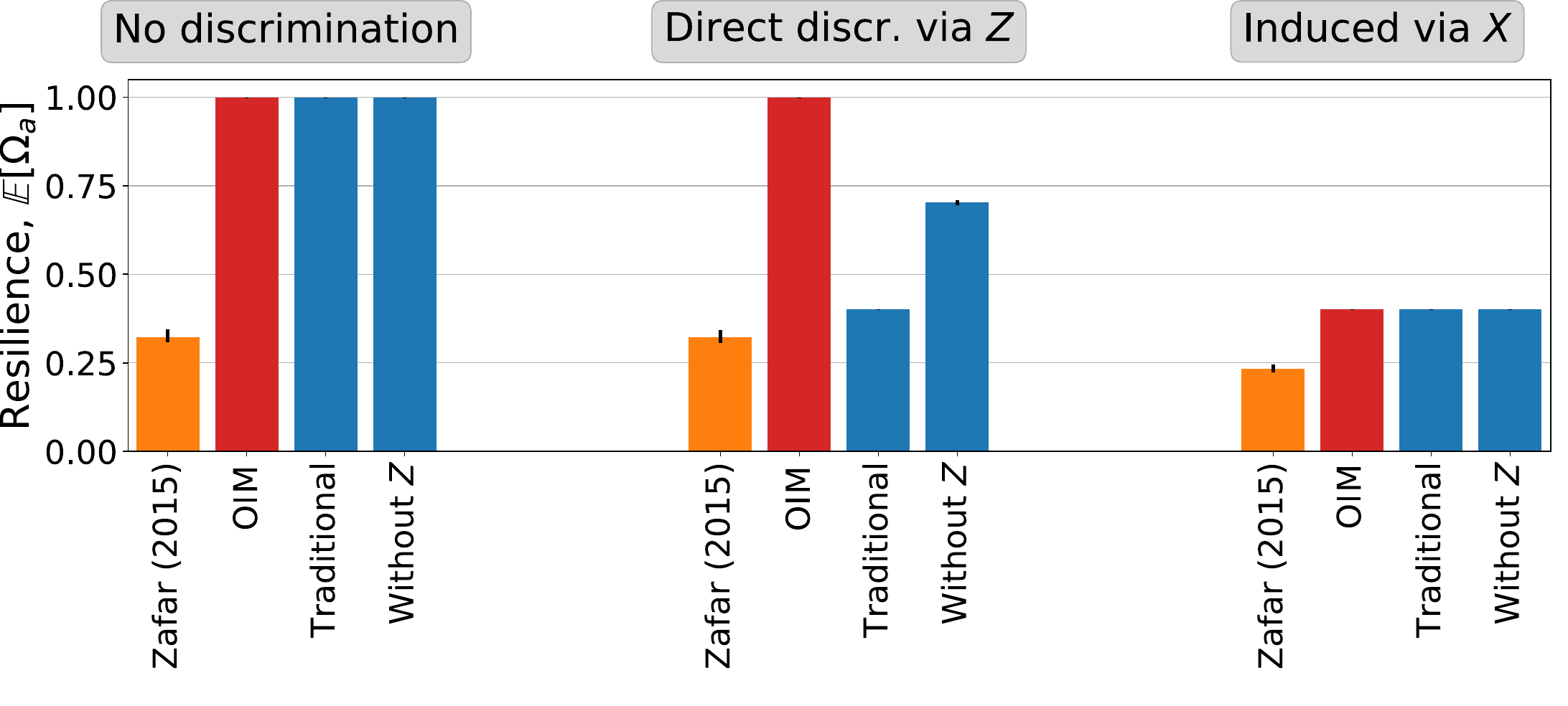}
    \caption{Linear regression}
  \end{subfigure}
\caption{
Resilience of various learning algorithms for logistic regression (upper) averaged over datasets. Error bars correspond to $95\%$ confidence intervals of the expectation, obtained via bootstrapping.
}
\label{fig:mini}
\end{figure}

%
%
We check whether the results from \S\ref{sec:synthetic} hold over various parameters of data generating processes. 
For each learning algorithm, the procedure of data generation and training is repeated $1000$ times, each time with a different correlation matrix $\Sigma$ and parameters \bm{$\alpha$}, \bm{$\tilde{\alpha}$}, $\beta$ (additional details in Appendix C). We report mean resilience of each learning algorithm, averaged over randomly generated datasets (Figure~\ref{fig:mini}).

When the learning algorithms preventing discrimination are applied to non-discriminatory data, they should fall back to a traditional learning algorithm to avoid biases in inference and yield perfect resilience. For logistic regression, only two algorithms achieve this for all datasets: the method based on envy-freeness (``Zafar EF'' in the upper leftmost Figure~\ref{fig:mini})~\cite{Zafar2017Parity} and our OIM (the red bar in the upper leftmost Figure~\ref{fig:mini}).
The OIM is also more resilient to directly discriminatory perturbations than other supervised methods aiming to prevent discrimination for logistic regression (the middle and rightmost panels in upper Figure~\ref{fig:mini}).
The second best method is traditional learning (with or without the protected attribute; blue bars in upper Figure~\ref{fig:mini}), and third is the game-theoretic method based on envy-freeness(``Zafar EF'' in upper Figure~\ref{fig:mini}). However, these two methods allow direct discrimination via $Z$ (middle upper Figure~\ref{fig:mini}). 

The difference between the OIM and traditional learning is small for logistic regression (upper Figure~\ref{fig:mini}), but it is large for linear regression (lower Figure~\ref{fig:mini}).
For the linear regression model, the proposed method achieves maximal resilience to directly discriminatory perturbations (lower Figure~\ref{fig:mini}). Here the difference in resilience between the OIM and other methods is significantly greater than that for logistic regression (upper Figure~\ref{fig:mini}).

\section*{Appendix F: Features affected by direct discrimination}
Apart from the perturbations of the output variable, $U$, the perturbed dataset, $\tilde{D}$ could also include the perturbations of some of the relevant attribute $X_1$. We refer to such relevant attribute as $\tilde{X}_1$. From the perspective of decisions $Y$, such perturbations result in indirect discrimination, because they impact $Y$ indirectly through $\X$.
For instance, Jim Crow laws required literacy to decide whether an individual has a voting right, while ethnic minorities had systematically limited access to education~\cite{klarman2006jim}.
%
If some $\tilde{X}_1$ is the outcome of human decisions and is affected by direct discrimination, then we could and should apply the same reasoning and methods as we do to $Y$, i.e., we shall construct a respective model for $\tilde{X}_1$, in which this variable is treated as an output variable. Then, one can obtain an estimator of $X_1$ based on $\tilde{X}_1$ by applying the OIM. The computed OIM of $X_1$ can be used to also obtain an estimator of $U$ based on $Y$. 
We apply this procedure within our evaluation framework by modeling a perturbation of $X_1$ in the same manner as of $\U$.
We measure the resilience of the learning algorithms to this perturbation finding that the OIM prevents direct discrimination in $\X$ and as a consequence in $U$ (Figure~\ref{fig:disc_x}), under a linear model of $\X$ and either a logistic or linear model of~$Y$.

\section*{Appendix G: Missing features.} 
In real-world settings, attributes are often unknown or their measurements are unavailable. We model this scenario by removing $X_2$ from the training dataset $\tilde{D}$, while keeping it unchanged in $D$. Then, we measure the resilience of learning algorithms to the non-discriminatory and discriminatory perturbation.

When $X_2$ is missing, we obtain nearly identical relative resilience results as before.
The OIM is more resilient to direct and induced discriminatory perturbations than the other supervised methods aiming to prevent discrimination (Figure~\ref{fig:missing}). For logistic regression, the game-theoretic method based on envy-freeness, ``Zafar EF'' (the upper middle and rightmost panels in Figure~\ref{fig:missing}) has only slightly worse performance for direct discrimination, and the same or slightly better performance when there is no discrimination or when there is induced discrimination. Since these methods are missing one of the attributes required to model the data generating process, their predictions are significantly worse and resilience is considerably less than in the case where all attributes are available for training. However, except for ``Zafar EF'', the resilience remains similar in ranking between methods to the scenario where all attributes are available.

\section*{Appendix H: Non-linear models.} 
In addition to missing features, real-world data may be generated by complex non-linear processes that cannot be fit using simple models like logistic regression.
To simulate this scenario, we introduce a non-linearity in $f(\x)$. Here we present the results for $f(\x)=\alpha_1 x_1 x_2$, but we obtain the same qualitative results for other functional forms, such as $f(\x)=\alpha_1 \exp(\alpha_2 x_1 x_2)$ and  $f(\x)=\alpha_1 \sin(\alpha_2 x_1 x_2)$, where parameters $\alpha_i$ are random as in Appendix C. To learn these more complex models, we apply the OIM to deep neural networks (OIM-NN). 
We utilize a relatively simple architecture: three-fully connected hidden layers with the ReLU activation function and a sigmoid output layer. 
The hyperparameters are tuned to optimize accuracy as usual. Most other methods do not have implementations for deep learning models, so we cannot evaluate them, except for the traditional learning and the post-processing method based on equalized odds~\cite{Hardt2016Equality}.

To provide more details, we report the cumulative distribution function of per-dataset resilience of each learning algorithm, $\Omega_a$, divided by the resilience of the OIM-NN, $\Omega_\text{OIM}$, for classification (Figure~\ref{fig:nn}). 
The deep learning models are more resilient to data perturbations than their logistic regression counterparts for nearly all datasets (``NN'' versus ``LR'' in Figure \ref{fig:nn}), since neural networks are better suited to approximate the non-linear data. 
Most importantly, the OIM-NN tends to outperform all other methods. When compared to the traditional deep learning without $Z$, the OIM-NN is more resilient to directly discriminatory perturbations of data for 80\% of datasets (blue dashed line in the middle Figure \ref{fig:nn}).

\section*{Appendix I: The choice of image markings}
We show that the results between an alternative box marking style (Figure \ref{fig:celebaa}a) and the box marking style presented in the main text (Figure \ref{fig:celeb} \& \ref{fig:celebsize}) are nearly identical. Furthermore with this alternative marking style, we show how the effect of the marker size affects the performance of the OIM as in the main text (Figure \ref{fig:celebsizesa}). 


\end{document}


\linenumbers

\section*{Appendix A} 


Here, we present the results from a synthetic scenario proposed by \cite{Lipton2018Does}, modified slightly as follows. Using this example, we show how state-of-the-art learning algorithms addressing discrimination induce it even when the training data is non-discriminatory. 

To this end, we sample 1000 observations from the data-generating process below:
\begin{gather*}
    z_i \sim \text{Bernoulli}[0.5]\\
    \text{hair\_length}_i | z_i = 0 \sim  35 * \text{Beta[2, 2]}\\
    \text{hair\_length}_i | z_i = 1 \sim  35 * \text{Beta[2, 7]}\\
    \text{work\_exp}_i | z_i = 0 \sim \text{Poisson}[25] - \text{Normal}[20, \sigma = 0.2]\\
    \text{work\_exp}_i | z_i = 1 \sim 
\begin{cases}
    \text{Normal}[10, \sigma = 2]& \text{w/ prob 0.2}\\
    \text{Normal}[15, \sigma = 2]& \text{w/ prob 0.8}
\end{cases} \\
    p_i= f(-25.5 + 2.5*\text{work\_exp}) \text{ where } f(x) = \frac{1}{1+e^{-x}} \\
    y_i | \text{work\_exp} \sim \text{Bernoulli}[p_i]
\end{gather*}
This synthetic data represents the historical hiring process where the protected attribute is a candidate's gender, $z$. The data has the following properties: i) the hiring decision has been made based on the work experience only, thus, it is non-discriminatory data;  ii) since women in this scenario on average have less work experience than men, men have been hired at higher rate than women historically; and iii) women tend to have longer hair than men. Therefore, a model that uses hair length in its decision-making can induce indirect discrimination. Additionally, we introduced modifications to this synthetic data with respect to the original scenario~\cite{Lipton2018Does}. The work experience of male candidates now follows a bi-modal distribution (i.e., a mixture of two normal distributions) with one peak at 10 and another at 15. We trained a method for discrimination prevention \cite{Zafar2017Fairness} under three different fairness constraints: equalized \textit{misclassification rate}, \textit{false positive rate} (FPR), \textit{false negative rate} (FNR). We also trained a model while simultaneously optimizing both FPR and FNR; however, the learned model returned trivial predictions where all candidates are rejected.
The relative utility of the various methods is low compared with the OIM.

\begin{table}[ht]
\begin{center}
\centering
\begin{tabular}{l | c} 
 Method & $\E[R_\text{perf}/R]$  \\
 \hline
 OIM& 1.000  \\ 
 \citet{Zafar2017Parity} & 0.997 \\
 \citet{Zafar2017Fairness} with FNR & 0.838 \\
 \citet{Zafar2017Fairness} with Missclass. & 0.777 \\
 \citet{Donini2018Empirical} & 0.634 \\
 \citet{Zafar2017Fairness} with FPR & 0.570  \\
 \citet{Hardt2016Equality} & 0.328 \\
 \citet{Zafar2015Fairness}& 0.179 \\
\end{tabular}
\par\bigskip
\par\bigskip
\caption{Relative utility of various fairness models \cite{Hardt2016Equality, Donini2018Empirical, Zafar2015Fairness, Zafar2017Fairness, Zafar2017Parity} trained with the synthetic data}
\label{tab:results}
\end{center}
\end{table}

%

\section*{Appendix B} %
\paragraph{Synthetic experiment setup.}
%
We report the performance of the model by Donini et al \cite{Donini2018Empirical} with SVM with linear kernel.
The regularization parameter $C$ was tuned via grid search with $C \in \{0.01,0.1,1\}$. 
We report the statistics of \cite{Zafar2017Fairness} when the model is optimized to equalize misclassification rates between two groups. The implementation of the models we used for the experiment \cite{Zafar2015Fairness, Zafar2017Fairness, Zafar2017Parity, Hardt2016Equality, Donini2018Empirical} are readily available online.

\begin{figure}[tb]
\centering
\includegraphics[width=0.48\textwidth]{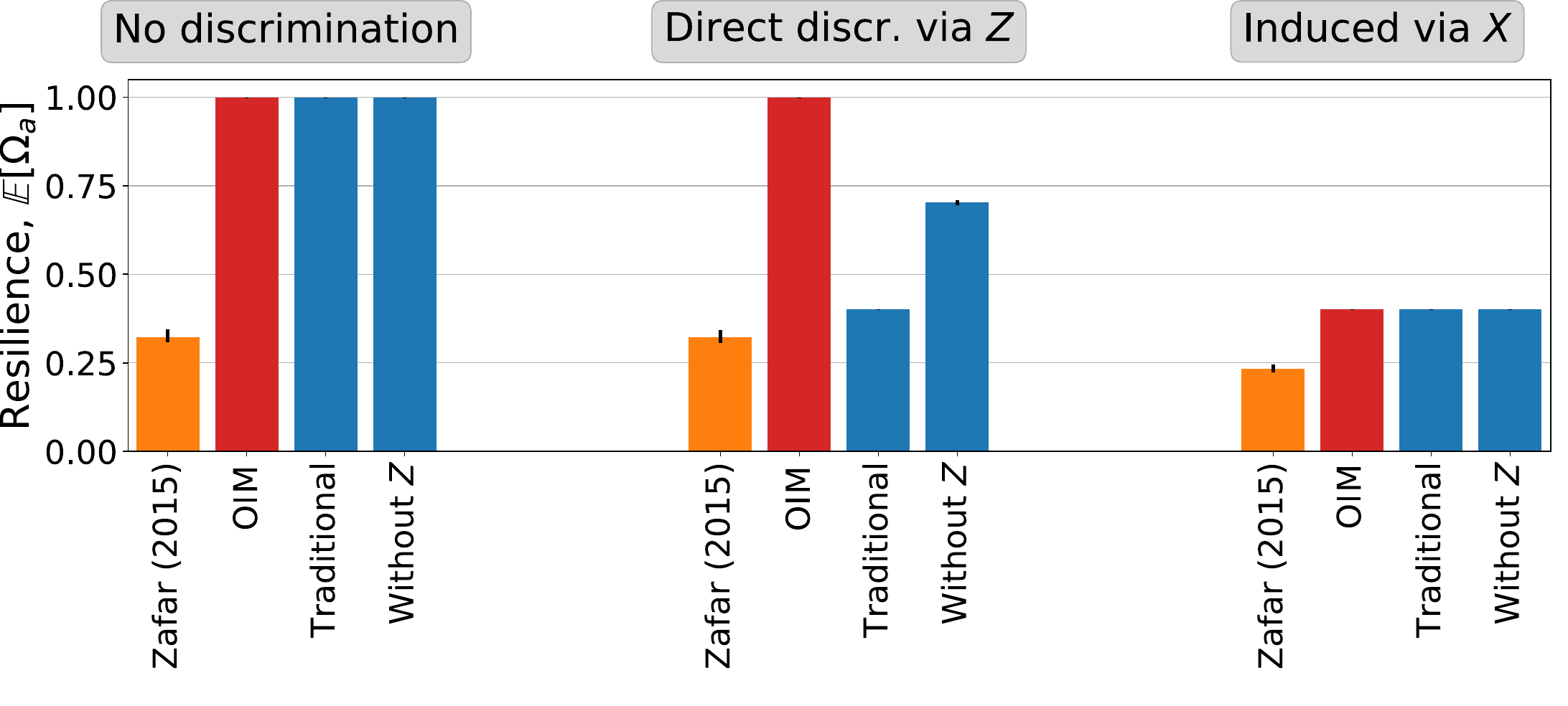} 
\caption{
Average resilience of learning algorithms to non-discriminatory perturbations (the leftmost column) and discriminatory perturbation (the remaining two columns), for linear regression models.
}
\label{fig:linearreg}
\end{figure}

\begin{figure*}[tb]
\centering
\includegraphics[width=0.8\textwidth]{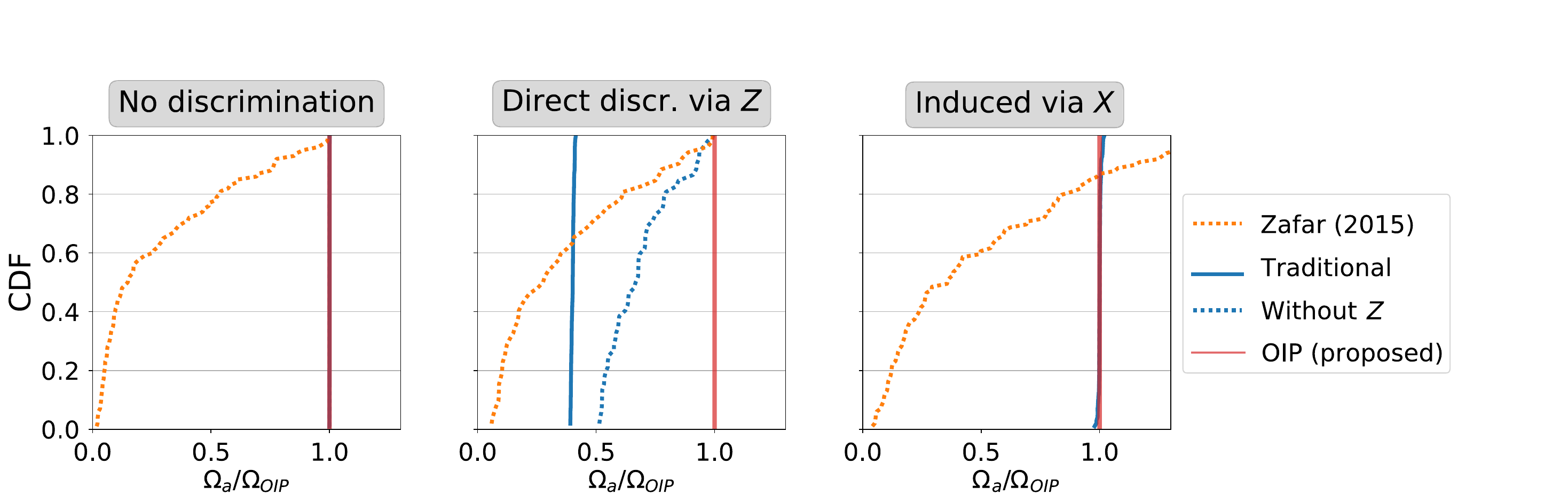}
\caption{
The cumulative distribution function of per-dataset resilience values divided by the resilience of the optimal imputing predictor for the linear regression dataset. The vertical red lines correspond to the optimal imputing predictors.
}
\label{fig:linearreg1}
\end{figure*}

\paragraph{Linear regression.} 
For the linear regression model, the proposed method achieves maximal resilience to directly discriminatory perturbations (see Figures \ref{fig:linearreg} \& \ref{fig:linearreg1}). Here the difference in resilience between the OIP and other methods is significantly greater than that for logistic regression (main text).

\section*{Appendix C}

\begin{figure}[htb]
\centering

\centering
\includegraphics[width=0.49\textwidth]{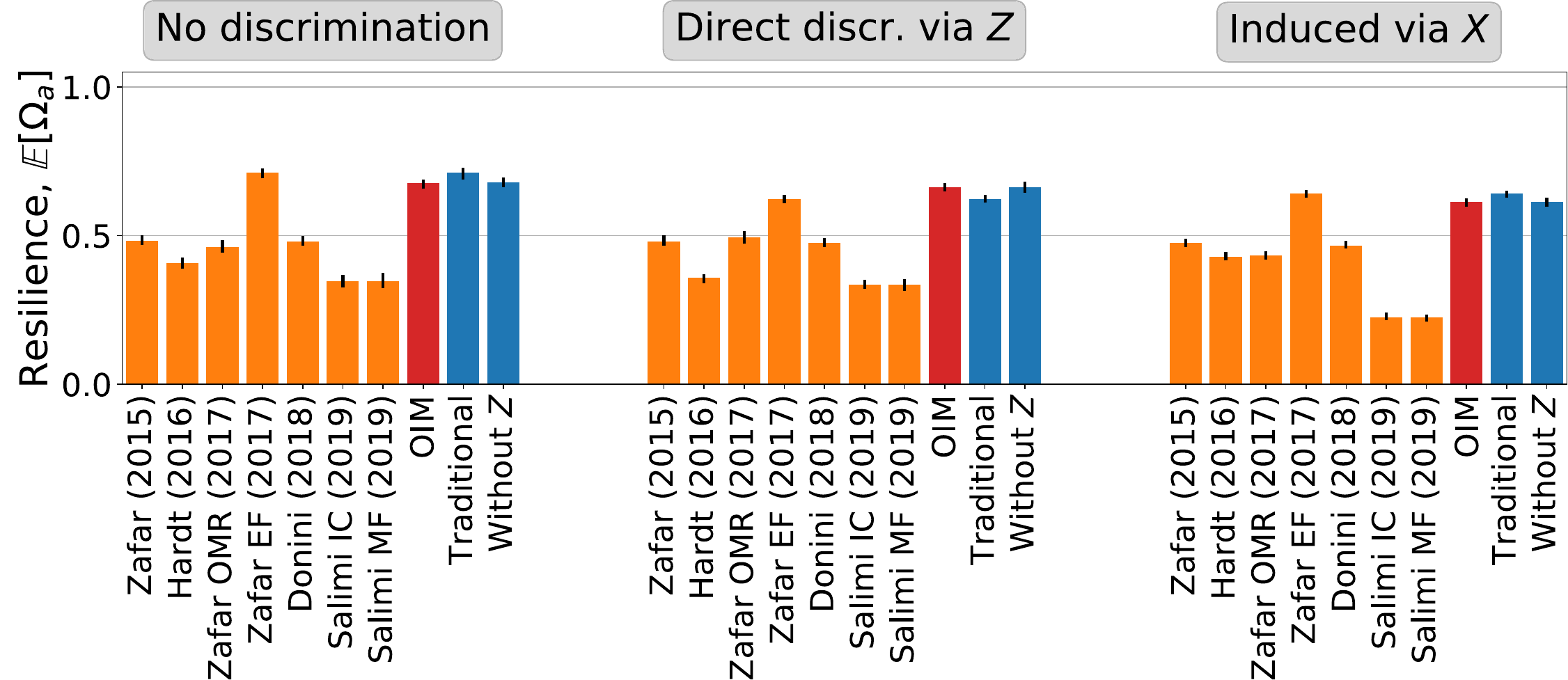} 
\includegraphics[width=0.49\textwidth]{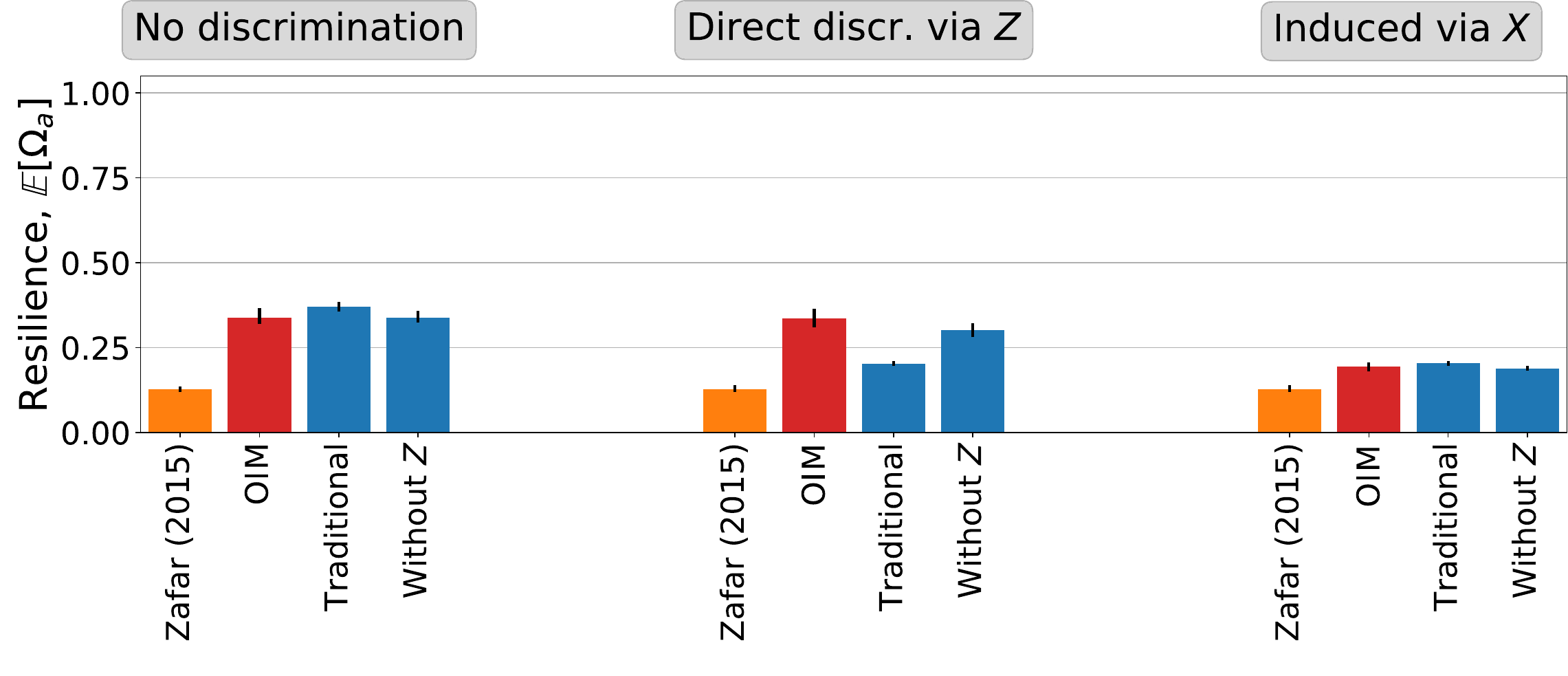} 
\caption{
Resilience of learning algorithms with $X_2$ missing during training to non-discriminatory perturbations (the leftmost column) and discriminatory perturbation (the remaining two columns), for logitic regression and linear regression.
}
\label{fig:missing}
\end{figure}

\paragraph{Missing attributes.}
In real-world settings, attributes are often unknown or their measurements are unavailable. We model this scenario by removing $X_2$ from the training dataset $\tilde{D}$, while keeping it unchanged in $D$. Then, we measure the resilience of learning algorithms to the non-discriminatory and discriminatory perturbation.
When $X_2$ is missing, we obtain nearly identical relative results as in the main text for direct and indirect discriminatory perturbations, when making relative comparisons among the learning methods.
The OIP is more resilient to direct and indirect discriminatory perturbations than the other supervised methods aiming to prevent discrimination. The game-theoretic method based on envy-freeness, ``Zafar EF'' (the middle and rightmost panels in Figure~\ref{fig:missing}) has only slightly worse performance for direct discrimination, and the same or slightly better performance when there is no discrimination or there is indirect discrimination. Since these methods are missing one of the attributes required to model the data generating process, their predictions are significantly worse and resilience is considerably less than in the case where all attributes are available for training. However, except for ``Zafar EF'', the resilience remains similar in ranking between methods to the scenario in the main text when all attributes are available.

%


\section*{Appendix D}
\begin{figure}[tb]
\centering
\includegraphics[width=0.7\textwidth]{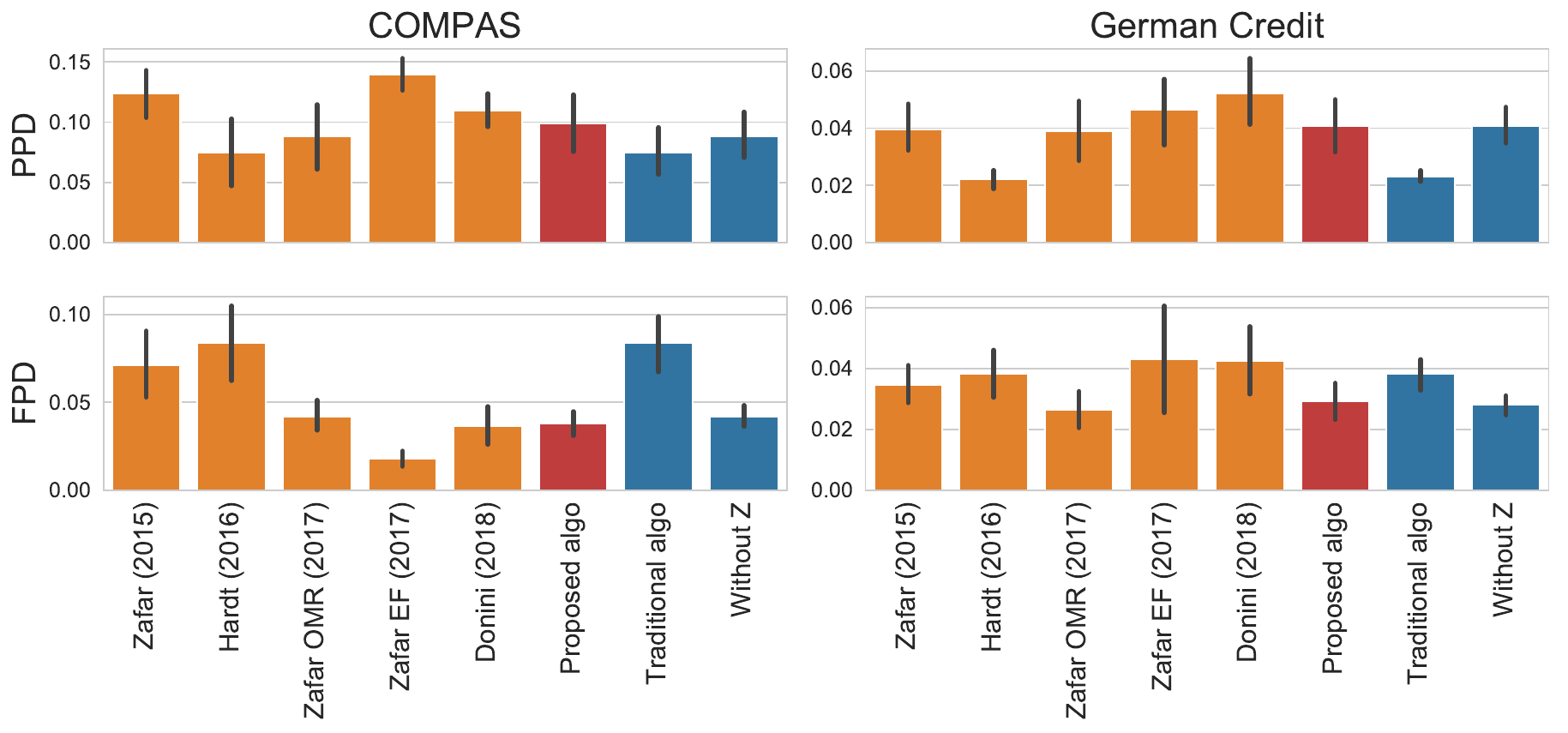}
\caption{Additional experiment with real-world dataset using Positive Predictive Disparity (PPD) and False Positive Disparity (FPD). The lower these values are, the more likely that models are fair.}
\label{fig:grid_appendix}
\end{figure}

\paragraph{Real-world experiment setup.}
Similar to the synthetic experiment, we report the performance of the model by Donini et al \cite{Donini2018Empirical} with a SVM and with a linear kernel. The regularization parameter $C$ was tuned via grid search with $C \in \{0.01,0.1,1\}$. We report the statistics of \cite{Zafar2017Fairness} when the model is optimized to equalize misclassification rates between two groups. The fairness metrics used in this experiment setup are described in Table \ref{tab:fairness_metric}. The results for Positive Predictive Disparity and False Positive Disparity, which are not in the main text, can be seen in Figure \ref{fig:grid_appendix}.

\paragraph{COMPAS Dataset.}
The ProPublica COMPAS dataset \cite{Larson2016How} contains the records of 7214 offenders in Broward County, Florida in 2013 and 2014. COMPAS also provides binary label for each data if the individual shows high sign of recidivism. We use the race (African American, Caucasian) as the sensitive features. This dataset also includes information about the severity of charge, the number of prior crimes, and the age of individuals. 

\begin{center}
\begin{table}
\centering
\begin{tabular}[tb]{||c||} 
\hline
 Description and Definition  \\
 \hline\hline
 Demographic Disparity (DD): \\ 
    $|P(\hat{y}=1|z=0) - P(\hat{y}=1|z=1)|$\\
    \hline
 Positive Predictive Disparity (PPD):\\
     $|P(y=1|\hat{y}=1, z=0) - P(y=1|\hat{y}=1, z=1)|$\\
     \hline
 False Positive Disparity (FPD):\\
     $|P(\hat{y}=1|y=0, z=0) - P(\hat{y}=1|y=0, z=1)|$\\
     \hline
\end{tabular}
\par\bigskip
\par\bigskip
\caption{Summary of discrimination metrics used in our real-world experiments}
\label{tab:fairness_metric}
\end{table}
\end{center}

\paragraph{German Credit Dataset.}
German Credit Dataset \cite{Dua:2019} provides information about 1000 individuals and the corresponding binary labels describing them as creditworthy ($y_i$= 1) or not ($y_i$= 0). Each feature $x_i$ includes 20 attributes with both continuous and categorical data. We use the gender of individuals as the sensitive feature. This dataset also includes information about the age, job type, housing type of applicants, the total amount in saving accounts, checking accounts and the total amount in credit, the duration in month and the purpose of loan applications.



\nobibliography{jabreftrim1,manual_additions}